\documentclass[conference]{IEEEtran}

\usepackage{microtype}
\usepackage{graphicx}
\usepackage{enumitem}

\usepackage{subcaption} 
\usepackage{booktabs} 
\usepackage{array} 
\usepackage{hyperref}

\newcolumntype{P}[1]{>{\centering\arraybackslash}p{#1}}
\usepackage{cite}

\ifCLASSINFOpdf
\else
\fi
\usepackage{amsmath}
\begin{document}
%
\title{Handwriting styles: benchmarks and evaluation metrics}

\author{\IEEEauthorblockN{Omar Mohammed}
\IEEEauthorblockA{
Univ. Grenoble-Alpes\\
GIPSA-lab, LIG-lab\\
38000 Grenoble, France\\
Omar-Samir.Mohammed@grenoble-inp.fr}
\and
\IEEEauthorblockN{Gerard Bailly}
\IEEEauthorblockA{
Univ. Grenoble-Alpes\\
GIPSA-lab\\
38000 Grenoble, France\\
Gerard.Bailly@grenoble-inp.fr}
\and
\IEEEauthorblockN{Damien Pellier}
\IEEEauthorblockA{
Univ. Grenoble-Alpes\\
LIG-lab\\
38000 Grenoble, France\\
damien.pellier@univ-grenoble-alpes.fr}
}

%


\maketitle

\begin{abstract}
Evaluating the style of handwriting generation is a challenging problem, since it is not well defined. It is a key component in order to develop in developing systems with more personalized experiences with humans. In this paper, we propose baseline benchmarks, in order to set anchors to estimate the relative quality of different handwriting style methods. This will be done using deep learning techniques, which have shown remarkable results in different machine learning tasks, learning classification, regression, and most relevant to our work, generating temporal sequences. We discuss the challenges associated with evaluating our methods, which is related to evaluation of generative models in general. We then propose evaluation metrics, which we find relevant to this problem, and we discuss how we evaluate the evaluation metrics. In this study, we use IRON-OFF dataset \cite{791823}. To the best of our knowledge, there is no work done before in generating handwriting (either in terms of methodology or the performance metrics), our in exploring styles using this dataset.
\end{abstract}

\section{Introduction}

\par The characterization and the extraction of human style profile, given some human activity (like speech, handwriting, human interactions,...etc), is an open research problem. Usually, there is no clear definition of styles, making style extraction an ill-posed problem. In case of generative models, taking styles into account allows us to have more personalized generation. 

\par In this paper, we look at the problem from the angle of generating for handwriting. Ideally, given a letter from a writer, we would like to have information about the letter symbol (the character) and the factors that give the characterize the shape (which, be default, characterize the writer). By doing so, we can: \textit{i}) better study what constitutes the human profile, and \textit{ii}) produce more human-acceptable samples.

\par In this paper, we discuss 4 methods to capture the style, to be used for biasing handwriting generation. The handwriting generation and two of the proposed style methods are based on state of the art in deep learning. We then propose our performance metrics, and the reasoning behind them. The cardinal power of style methods are known beforehand. This will allow us to validate our choice of the evaluation metrics. 

\section{Related work}
\par Some of the remarkable recent advances in deep learning \cite{Goodfellow-et-al-2016} happened in the area of generative models. For static data, such as generating images, the work done using \textit{Variational Autoencoders}~\cite{kingma2013auto} and \textit{Generative Adversarial Networks}~\cite{goodfellow2014generative} has shown remarkable results. 

\par In contrast, handling temporal data, such generating tracings from letter/writer embedding, is more challenging: the data is sequential, and it is difficult to keep the coherence for long sequences. Advances in neural networks cell structure, like \textit{LSTM} \cite{hochreiter1997long} and \textit{GRU} \cite{cho2014learning, chung2014empirical}, showed impressive results on handling long term dependencies in temporal sequences. 

\par These advances later allowed the development of state-of-the-art neural networks architectures for generating biased temporal sequences, specifically for generating text and image captioning, are showing impressive results \cite{sutskever2011generating, Sutskever:2014:SSL:2969033.2969173, karpathy2015deep, vinyals2015show}. More applications since then have been explored, like music generation \cite{briot2017music} and speech synthesis generation \cite{oord2016wavenet}.

\par The generation of continuous data has always been tricky. Graves \cite{DBLP:journals/corr/Graves13} combined \textit{Long-Short Term Memory} (LSTM) networks with \textit{Mixture Density Networks}, MDN \cite{bishop1994mixture}, to generate continuous handwritten characters, using \textit{IAM Handwriting Database} \cite{marti1999full}. While the results are impressive, the MDN approach are quite difficult to train. Another possible approach for generating continuous tracings is \textit{Gaussian Scale Mixtures}, GSM \cite{Theis:2015:GIM:2969442.2969455}. 

\par In order to simplify the procedure, and focus on our investigation of styles, we discretized the tracings using \textit{Freeman codes} for direction, and speed - see Section \ref{sec:preprocessing} more details -, and apply \textit{softmax} to the output of the last layer, instead of MDN. This was inspired by the results reported in \cite{VanDenOord:2016:PRN:3045390.3045575,oord2016wavenet}, where they show impressive results on the discrete domain, given a good discretization policy. Having a categorical distribution is more flexible and generic that a continuous distribution, and requires no assumption about the data distribution shape. 

\par Recently, interesting work has been done concerning style extraction in the area of speech synthesis \cite{DBLP:journals/corr/abs-1803-09047, DBLP:journals/corr/abs-1803-09017}. In their work, they extract a number of \textit{style tokens}. They evaluated the performance of their system via classical subjective rating of voice, and show these token relate to some aspects of the speech prosody and the speaker's voice.

\par On the evaluation side, there has been a lot of advancement in developing performance metrics for image captioning and machine translation \cite{Koehn:2010:SMT:1734086}. Metrics like \textit{BLEU}\cite{papineni2002bleu}, \textit{METEOR} \cite{banerjee2005meteor} and \textit{CIREr} \cite{vedantam2015cider} are considered the SOTA in image captioning and machine translation evaluation. Traditionally, the evaluation of these kind of applications is subjective. But with the advance of machine and statistical learning, there was a need to develop metrics that are cheap to evaluate, yet have a good correlation with the human evaluation.

\section{Dataset and Pre-Processing}

\subsection{Dataset} \label{sec:data}
\par
\begin{figure}[!htbp]
    \centering
    \includegraphics[scale=0.2]{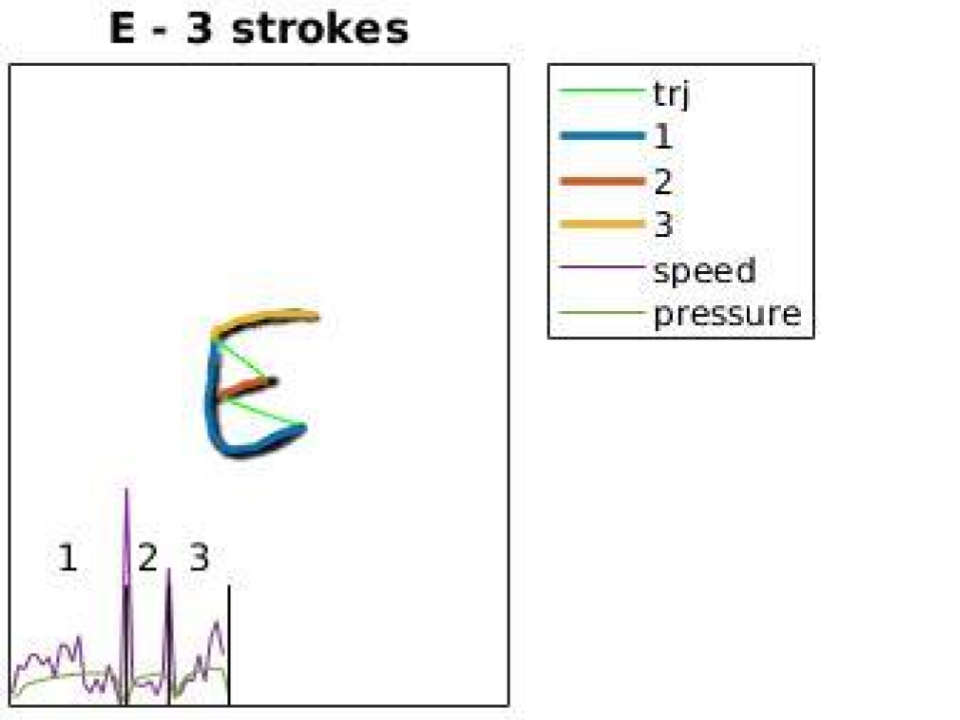}
    \caption{Example of a letter, showing the trajectory, strokes, speed and pen pressure}
    \label{fig:letter_example}
\end{figure}

 The dataset we choose is \textit{IRON-OFF} Cursive Handwriting Dataset \cite{791823}. While there are more famous handwriting datasets, like \textit{IAM Handwriting Database} \cite{marti1999full}, already available, our dataset provides us with separated and labeled letters, instead of entire sentence, thus allowing us to focus more on the problem of styles. A quick summary of this dataset is given below:

\begin{itemize}[noitemsep]
    \item ~700 writers total. We use 412 writers, who have written isolated letters.
    \item 10,685 isolated lower case letters.
    \item 10,679 isolated upper case letters, e.g. see Fig~\ref{fig:letter_example}.
    \item 410 euro signs.
    \item 4,086 isolated digits.
    \item Gender, handiness, age and nationality are available for all writers.
    \item For each letter, we have letter image - with size around ~167x214 pixels, and a resolution of 300 dpi -, pen movement timed sequence comprising continuous X, Y and pen pressure, and also discrete pen state. This data is sampled at 100 points per seconds on a Wacom UltraPad A4. 
\end{itemize}

\par One particular challenge in this dataset is that each writer wrote the letters only once. Since we are focusing on the styles, this makes it particularly challenging for us. We do not use the pressure or the pen state, in order to simplify the model.


\subsection{Pre-processing} \label{sec:preprocessing}

\par All the images of letters have been denoised and cropped in order to focus on the letters. Then, the images had been down-scaled to 28x28 pixels.

\par We cleaned the selected motion captured isolated letters by removing frames related to false starts or corrections, extra strokes as well as removing  entire  tracings  whose lengths exceed 1 second, in particular due to lengthy pen-up durations. All tracings exceeding 99 time steps has been discarded from the dataset as well. 

\par All the letter tracings are represented as two modalities: Freeman code - see part \ref{sub:freeman} - and speed features. Each modality is quantized into 16 level, and then represented as one-hot encoded vectors.

\subsubsection{Freeman coding for direction and quantizing speed} \label{sub:freeman}
Freeman codes~\cite{freeman1961encoding} belongs to a family of compression algorithms called Chain codes. These algorithms are useful to encode an image when it has connected components inside it. They are considered compression algorithms as they can transform a sparse matrix, to just a small fraction of the size of the image, in the form of a sequence of codes.
Original Freeman codes have 2 versions, 4-directional codes, and 8-directional codes. Both are fairly simple as they encode each direction with a unique number (from 0 to n-1, where n are the directions). A direction is defined in the image as the directed vector connecting two neighbouring pixels on the contour of a connected component.

\begin{figure}[htbp!]
\centering
\includegraphics[scale=0.4]{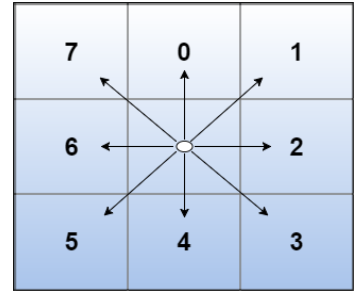}
\caption{Example for freeman code representation for 8 directions. Each direction will be given one number.}
\label{fig:freeman_dir}
\end{figure}

In our work, we compute the direction angle between each two consequent points. Then, we convert each direction to its corresponding freeman code symbol, as shown in Fig~\ref{fig:freeman_dir}. Then, we perform one-hot encoding on the direction, and feed it to our network. In order to have a faithful reconstruction of the letters, we also quantize the speed of each displacement.

\section{Models}
\subsection{Model selection}
The quality of generation of our model has been quite challenging -- due to the issues mentioned in the section~\ref{sec:data}. We ran random hyper-parameter selection for several days to get the best results.
The resulting generator is based GRU cell, with 3 hidden layers, each of size $256$, and a dropout of $0.3$. Adam optimizer \cite{kingma2014adam} is selected, with a learning rate of $10^-3$. An MLP is applied to the output of the GRU at each time step, with an output size of 34. Two softmax operation are then applied, one for output $1...17$, representing freeman codes, and the other on output $15...34$, representing the speed.
\par For the models used to extract styles/bias our generator, we followed a more conservative approach, starting from already tested architectures, and modifying their hyper-parameters gradually, until we got satisfying results. The architectures are reported in the following sections.

\subsection{Training}
\par
\begin{figure}[htpb!]
    \centering
    \begin{subfigure}[a]{0.9\columnwidth}
        \includegraphics[width=0.8\textwidth]{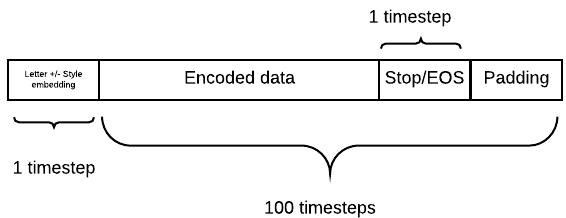}
        \caption{Input sequence shape}
        \label{fig:input_frame_shape}
    \end{subfigure}
    \begin{subfigure}[b]{0.9\columnwidth}
        \includegraphics[width=0.8\textwidth]{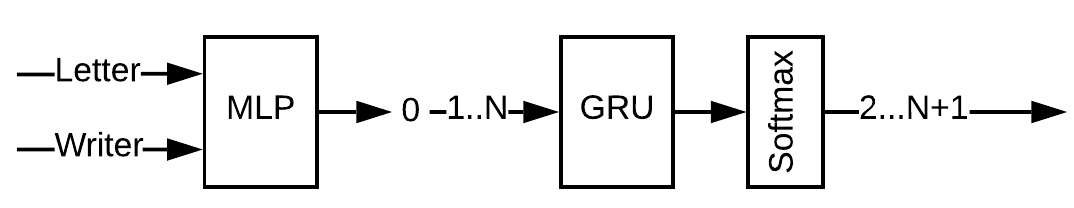}
        \caption{Training mode}
        \label{fig:training_mode}
    \end{subfigure}
    
    \begin{subfigure}[c]{0.9\columnwidth}
        \includegraphics[width=0.8\textwidth]{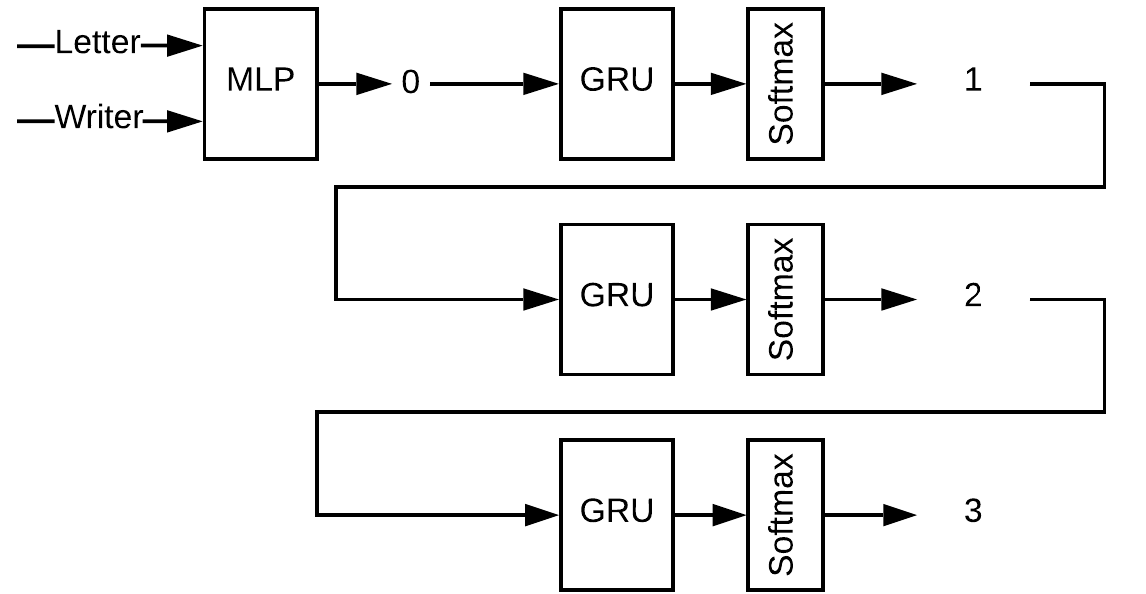}
        \caption{Generation framework}
        \label{fig:gen_mode}
    \end{subfigure}
    \caption{Illustration for biasing the generative model using letter + writer. The MLP, receives letter/writer embedding is responsible for down-/up-sizing the input dimension to the frame dimension of the tracings (34 i.e 16+1 hot encoding for direction and speed together with EOS feature). In this example, the model is biased using the letter and writer code.}\label{fig:model_details}
\end{figure}

 We follow the similar approach to the work done in image captioning~\cite{vinyals2015show}. Each handwritten character is encoded as shown in Fig~\ref{fig:input_frame_shape}. An \textit{End-of-Sequence} (EOS) symbol is added at the end of each sequence. Padding is done to make all sequence lengths equal. 

\par The first time step represents the bias we use for the model. It is projected to the same dimension as the rest of the letter sequence. For example, if we use the letter as embedding (as one-hot encoding, it has 26 dimensions), and dimensions of our sequence is 34 (16 + 1 for direction + EOS, 16 + 1 for speed + EOS), then we use a \textit{Multi-Layer perceptron} (MLP) to project the 26 dimensions into 34 dimensions

\par In the training phase, Fig~\ref{fig:training_mode}, first, a token that encodes the letter and the writer or his/her style is first set with the same feature dimension as the encoded sequence and considered as frame 0. This frame is added to rest of the encoded sequence (frames 1 to N) in order to bias the hidden states of the network. The objective of the model is to predict the next frame in the sequence given the preceding ones. The input to the model during the training is always the ground truth.

\par To formalize this, $S = (s_{1}...s_{t}) , t \in \{1...N\}$ is the input letter trace, where $N$ is the trace length, and $I$ is the letter with/without style - the model bias, and $Embedding$ is the MLP used to project $I$ to the same dimension as $S$, then our system works as the following:
\begin{equation}
    s_{0} = Embedding(I)
\end{equation}
\begin{equation}
    p_{t+1} = GRU(s_{t}), t \in \{1...N\}
\end{equation}

\par The loss used to optimized the GRU parameters is the negative log likelihood of the correct trace point at each time step, calculated as follows:
\begin{equation}
    L(S, I) = - \sum_{t=1}^{N} \log p_{t}(s_t)
\end{equation}

\subsection{Inference}
\par During inference, Fig~\ref{fig:gen_mode}, the first time step has the embedding information, used to bias the model.  The network then generates the first frame. This frame is then feedback to the network's input for generating the second frame. This continues until an EOS symbol is generated. 

\par Over the course of generation, the model accumulate errors, leading to degradation of performance when generating long sequences. Some techniques, like \textit{Scheduled Sampling} \cite{Bengio:2015:SSS:2969239.2969370}, can be applied during the training phase in order to enhance the quality of the model training, but they are not used in this work.

\par In order to infer/generate the tracing of the letter, we use the \textit{Softmax Sampling} strategy: at each time step, we generate a two multinomial distributions: one for the directions, the other for the speed). At time step $t$, we sample both distributions according to a \textit{temperature} level, and use these samples to feed the model's input for the next time step $t+1$. This method is the one we use in this work. 

\section{Biasing the network with a style input}

\par
\begin{figure}[!htbp]
    \centering
    \begin{subfigure}{0.2\textwidth}
        \includegraphics[scale=0.6]{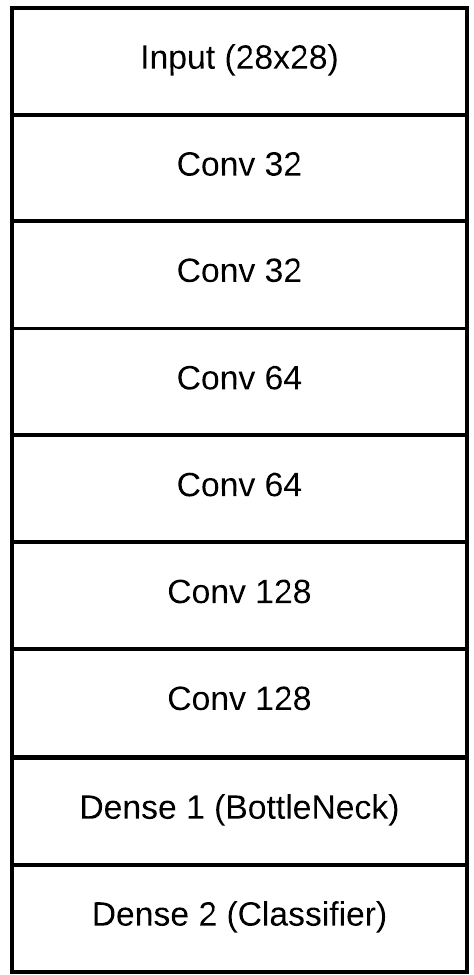}
    \end{subfigure}
    ~
    \begin{subfigure}{0.2\textwidth}
        \includegraphics[scale=0.6]{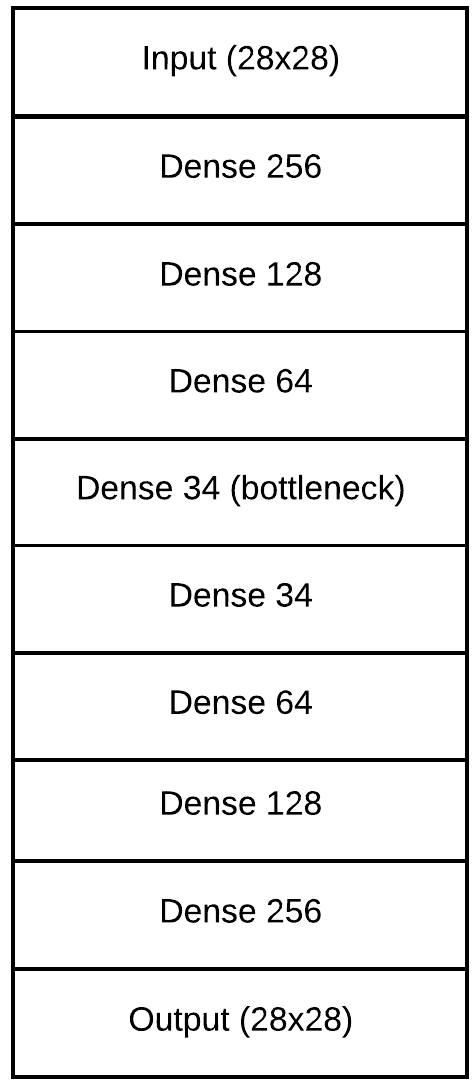}
    \end{subfigure}
    \caption{Left: architecture of the CNN letter classifier. Batch normalization is used after each convolution layer. The \textit{Dense 1} layer is the embedding that is used to bias our generator. Right: the autoencoder architecture we used. The first \textit{Dense 34} layer provides the latent space used to bias the generator.}
    \label{fig:architectures}
\end{figure}

 We assess the multiple methods to bias our letter generator in their ability to capture of writing styles. These methods are chosen since we know beforehand their cardinal order (which has more information than which). Knowing this information beforehand, we use it to ground our performance metrics. The methods are:
\begin{description}[noitemsep]
    \item[Letter bias]: the letter code is used as bias. No style information is thus included. We use this as a lower baseline.
    \item[Letter + Writer bias]: the letter and writer codes are used as bias. Thus, the model has an access explicit information about the writer (i.e. via his/her identity). Thus, this method is expected to perform the best. This model will also serve as a upper baseline.
    \item[Image classifier embedding] We build a convolution neural network (CNN) to classify the letters images, as shown in Fig~\ref{fig:architectures}. Our architecture achieves $95.1\%$ classification accuracy. The embedding layer will encode information about the discriminative distance between the letters. This model should perform the same or a bit less performance that the \textit{Letter bias}, since it learns to clusters the letters, and there are classification errors.
    \item[Image auto-encoder latent space] we train a letter image autoencoder, using reconstruction error, and use the latent space as a representation of the letter+style bias. The architecture we use can be seen in Fig~\ref{fig:architectures}. The latent space encodes the similarity between the letters. This model should perform worse than \textit{Letter bias}, since, while it capture the similarity between the letter images, it does not capture discriminative features about each letter.
\end{description}


\section{Experiments}

\subsection{Model selection}
The \emph{generator} part of our model has been quite challenging -- due to the issues mentioned in the \emph{Dataset and pre-processing} subsection. We ran random hyper-parameter selection for several days to get the best results.
For other models, used to extract styles/bias our generator, we followed a more conservative approach, where we started from architectures tested before, and modified their hyper-parameters gradually, till we got satisfying results.

\par Evaluation, in generative models, is by far the most challenging part. Ideally, we want metrics to capture the distance between the generated and the reference distributions of handwriting features, and not between images using an ink-deposition model~\cite{nguyen2010techniques}. In order to objectively compare the proposed style embeddings, we propose the following metrics:
\begin{description}
    \item[BLEU score~\cite{papineni2002bleu}] It is an important metric evaluate the quality of text generation areas, like in machine translation \cite{Sutskever:2014:SSL:2969033.2969173} and image captioning \cite{vinyals2015show}. In this work, we test the hypothesis that the BLEU score is also relevant to the generation of handwriting\footnote{In text evaluation, while the BLEU score is usually used when there is multiple reference sentences, there is no constraint on using it with one reference sentence only.}. In this study, we report the BLEU scores for 1, 2 and 3 grams, for the freeman codes and the speed separately.
    The final score is calculated as follows:
    \begin{equation}
    BLEU_{NG} = \frac{\sum_{C\in G}\sum_{NG\in C}Count_{Clipped}(NG)}{\sum_{C\in G}\sum_{NG\in C}Count(NG)}
    \end{equation}
    \begin{equation}
    Score_{N} = \min{(0, 1 - \frac{L_{R}}{L_{G}})} \prod^{N}_{x=1}BLEU_{N}
    \end{equation}
    where: $G$ is all the generated sequences, $NG$ is the N-gram to be measured, $N$ is the total number of N-grams we want to consider, $Count_{Clipped}$ is clipped N-grams count (if the number of N-grams in the generate sequence is larger than the reference sequence, the count is limited to the number in the reference sequence only), $L_R$ is the length of the reference sequence, $L_G$ is the length of the generated sequence. 
    The term $\min(0, 1 - \frac{L_{R}}{L_{G}})$ is added in order not to penalize small generated sequences (smaller than the reference sequence), which will achieve high scores.
    
    \item[Generated Sequence Length] Another aspect that we measure, is the relationship between the length of the generated sequence and the reference sequence. Thus, for each proposed method, we use the \textit{Wilcoxon signed-rank test}~\cite{10.2307/3001968} to compute the statistical significance between the distribution of the length of generated letters and the reference letters. In addition, we also calculate the~\textit{Pearson correlation coefficient} on the length as well, in order to better quantify the relation between the generated and the ground-truth letters.
\end{description}

\section{Results and Discussion}

\subsection{BLEU scores}
\par
\begin{figure*}
    \centering
    \begin{subfigure}[b]{0.48\textwidth}
        \includegraphics[width=\textwidth]{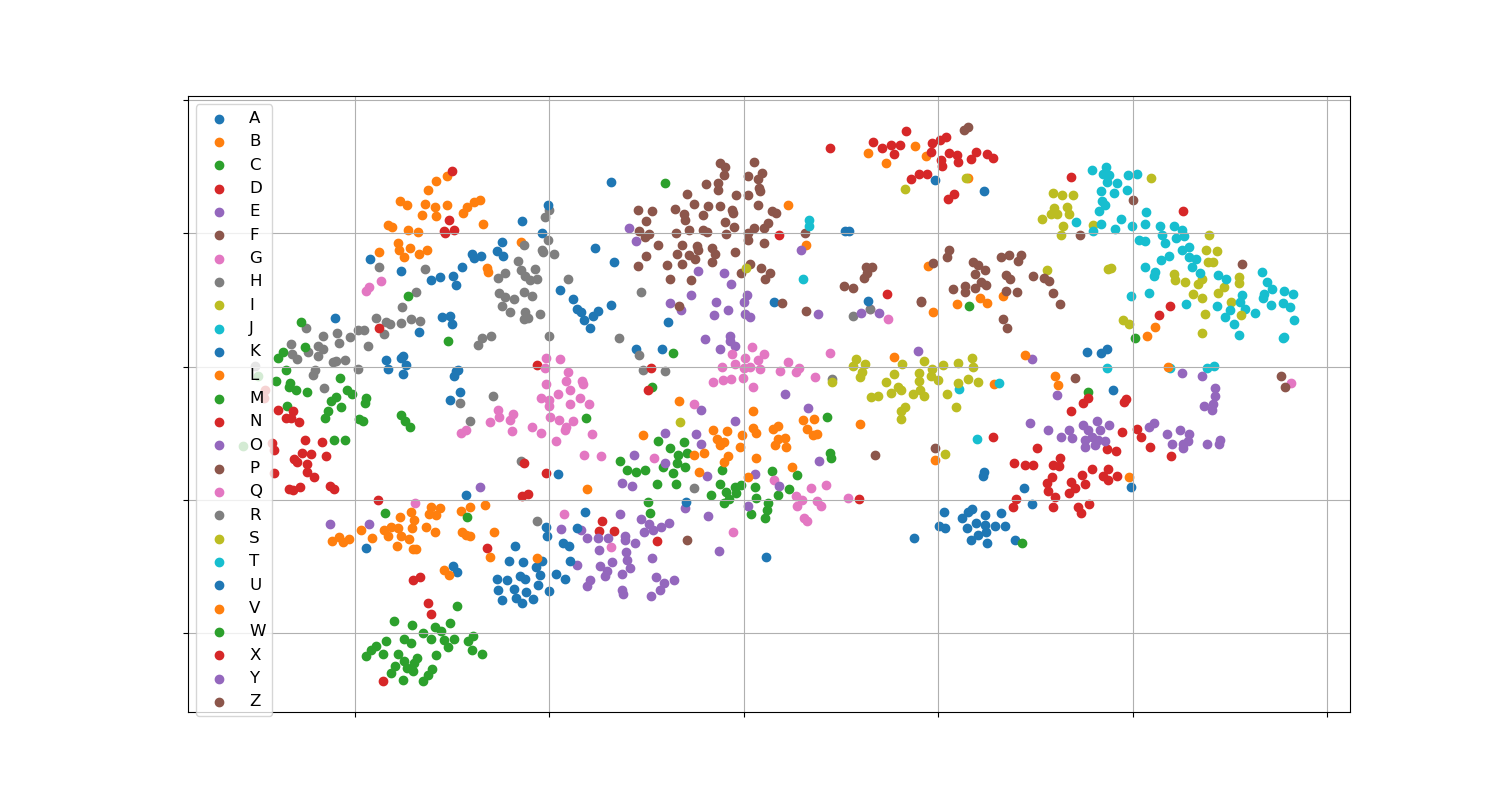}
        \caption{}
        \label{fig:autoenc_latent}
    \end{subfigure}
    \quad
    \begin{subfigure}[b]{0.48\textwidth}
        \includegraphics[width=\textwidth]{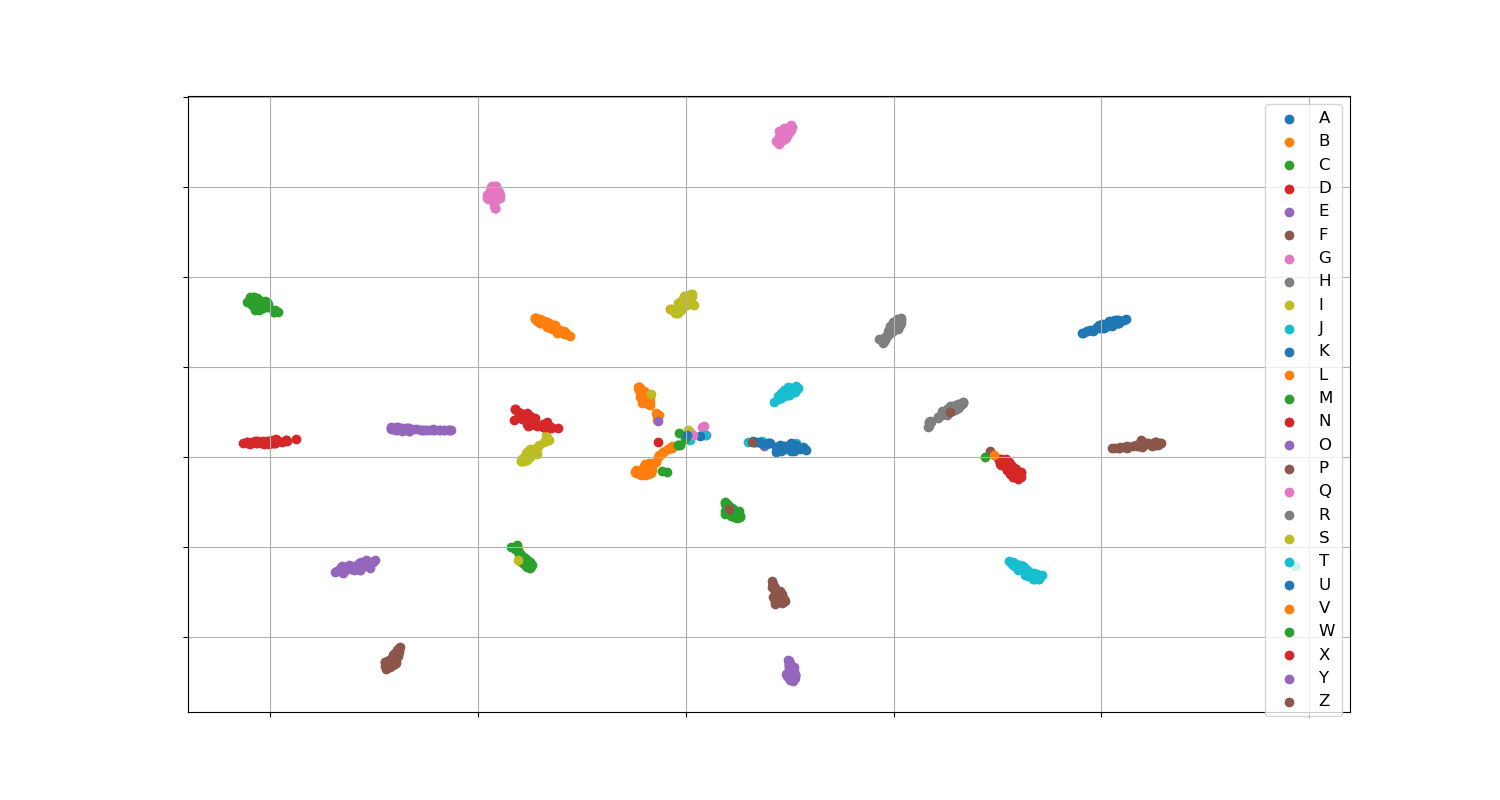}
        \caption{}
        \label{fig:classifier_latent}
    \end{subfigure}
    \caption{a) In the the autoencoder latent space, there is no clear separation between letters; the encoding is based on the similarity of the images only. b) In the classifier embedding, there is a clear separation between the letters - with few exceptions -.}
    
\end{figure*}

 The final results using the BLEU score can be seen in table \ref{table:1}. The following is observed:
\begin{itemize}
    \item The \textit{letter + writer} bias performed better than all other biases (in terms of B-3, for both speed and freeman directions), thus showing that having access to information about the writer, even so basic like the writer ID, have a clear advantage in the resulting quality of the handwriting generation.
    \item The embedding from the image autoencoder performed the worse. To understand why, we show a 2-D projection of the latent space using t-SNE in Fig~\ref{fig:autoenc_latent}. Since the autoencoder is trained for minimizing the reconstruction error only, the distance in the latent space encode mostly the proximity between the images with no distinct representations for letter and style. It can be seen that the model latent space doesn't encode discriminative features for the letters. Using this latent space for our generator, we find the model gets easily confused between nearby letters, leading to generating different letters than requested.
    \item The embedding from the image classifier performs better than the \textit{letter only} baseline, but the results vary compared to the \textit{letter+writer} model. Since the classifier is trained on a single objective only (to classify the letters), and the classifier performs very well, we can expect the embedding to cluster the letters well, as seen in Fig~\ref{fig:classifier_latent}. Also, we can expect the model to capture some of the writer style, possibly in the inter-cluster variance. This is an interesting result, suggesting that some fine tuning for the image classifier while in the generation task could be beneficial.
\end{itemize}

\begin{table*}[!htbp]
\centering
\begin{tabular}{|P{3cm}||P{1.5cm}|P{1.5cm}|P{1.5cm}||P{1.5cm}|P{1.5cm}|P{1.5cm}|} 
\hline
\multicolumn{1}{|c||}{Aspect/Feature} & \multicolumn{3}{c||}{ Speed } & \multicolumn{3}{c|}{ Freeman }   \\
\hline
Model / B-score      & B-1  & B-2  & B-3           & B-1  & B-2   & B-3              \\ \hline
Letter bias          & 49.7 & 37.3 & 24.2          & 47.4 & 36.6  & 26.8               \\\hline
Image classifier     & 50.9 & 38.2 & 24.6          & 48.5 & 37.9 & 28.1             \\\hline
Image autoencoder    & 51.9 & 37.9 & 23.1          & 46.4 & 35.0  & 24.5             \\\hline
Letter + Writer bias & 51.5 & 41.4 & 25.1          & 56.7 & 39.4  & 28.3             \\\hline
\end{tabular}
\caption{Comparing different approaches for style extraction using clipped n-grams}
\label{table:1}
\end{table*}

\subsection{EOS performance}
\par As mentioned earlier, we performed a statistical test between the paired distributions of lengths of the generated and the reference letters -- in other words, when the EOS symbol first appears. The results are shown in table~\ref{table:2}. We can see the following: 
\begin{itemize}
    \item For the statistical test, we can see that \textit{letter+writer} bias outperform the rest of the approaches, achieving p-value $< 0.05$. This is quite reassuring, since it is also in line with the results from the BLEU score.
    \item The results from the Pearson correlation coefficients are also consistent with the rest of the results. High coefficients are given to the \textit{letter+writer} biases, compared to the other methods. The image classifier and autoencoder gives the lowest results. This can be due to the errors during the learning, and the insufficient information about the letter length that can be inferred from the image. For the image classifier, as noted earlier, a fine-tuning during the generation task is worth exploring.
\end{itemize}

\begin{table}[!htbp]
\centering
\begin{tabular}{|c|c|c|}
\hline
Models & Pearson coefficient & p value \\ \hline
Letter bias & 0.38 & 0.84 \\ \hline
Image classifier & 0.32 & 0.62\\ \hline
Image autoencoder & 0.25 & 0.29 \\ \hline
Letter + Writer bias & 0.55 & 0.04\\ \hline
\end{tabular}
\caption{Pearson correlation coefficients and associated p-values for the EOS distributions of the different style biases.}
\label{table:2}
\end{table}

\section{Conclusions and future work}

\par We have proposed baselines for the task of handwriting generation, and evaluation metrics in order to measure the quality of the different methods: a \textit{letter} bias only, which capture the average of the letters, and a \textit{letter + writer} bias, which has a direct access to the writer ID (and thus, has information about the style). We also proposed two performance metrics: BLEU score (adapted from machine translation) and EOS analysis. In order to ground those metrics, we leveraged our prior knowledge over the cardinal power of different styling methods. With the performance metrics matching our expectation, we show a logical argument for using this metrics in the future for this task. This is an essential first step, towards further study and analysis for styles in handwriting, enabling further techniques to be developed and compared to each other.

\par Multiple points can be done in order to enhance our results, or to extend our study to become more complete. For example:
\begin{description}
    \item[Extract styles from examples]: The \textbf{lettter + writer} bias has explicit access to the writer ID, which we argue is the simplicity possible style information about the writer. The advantage is that it is quite simple, yet it does not have much information about the writer. For examples, for the \textit{X} letter, some people draw it clockwise and some anticlockwise. Some people start from the left side, and some started from the right side.
    \item[Style transfer]: From our observation of the data, although there are ~400 writers, there are some components for writing styles, like the ones mentioned in the letter \textit{X} in the previous point (although it is not possible to enumerate them). One way to test the quality of a style extraction method is by performing a style transfer: leveraging the information from different writers to make a quick adaption to a new unseen writer. One interesting method we are investigating at the moment to extract the writer style is to adapt the method used in \textit{FaceNet} \cite{DBLP:journals/corr/SchroffKP15}, where they want to create an embedding for human faces. They introduced a loss function, \textit{the triplet loss}, which is generic enough to be used in other applications, like identifying the speaker turn \cite{DBLP:journals/corr/Bredin16}. Also, recent work has been performed in style transfer in the domain of speech synthesis\cite{DBLP:journals/corr/abs-1803-09047, DBLP:journals/corr/abs-1803-09017} for separating textual input from voice and expressivity shows promising results.
    \item[Task specific metrics]: The proposed metrics in this paper are quite generic, allowing us to evaluate the system as a whole. Yet, a better understanding and analysis for the different systems requires more task-specific metrics. This is also in-line with the previous points, since it will give better insight on developing better methods for writer style extraction.
\end{description}


\section*{Acknowledgment}
This work is supported by PERSYVAL (ANR-11-LABX-0025) via the project-action RHUM.



%



\bibliographystyle{IEEEtran}
\bibliography{bibliography}

\begin{thebibliography}{10}
\providecommand{\url}[1]{#1}
\csname url@samestyle\endcsname
\providecommand{\newblock}{\relax}
\providecommand{\bibinfo}[2]{#2}
\providecommand{\BIBentrySTDinterwordspacing}{\spaceskip=0pt\relax}
\providecommand{\BIBentryALTinterwordstretchfactor}{4}
\providecommand{\BIBentryALTinterwordspacing}{\spaceskip=\fontdimen2\font plus
\BIBentryALTinterwordstretchfactor\fontdimen3\font minus
  \fontdimen4\font\relax}
\providecommand{\BIBforeignlanguage}[2]{{%
\expandafter\ifx\csname l@#1\endcsname\relax
\typeout{** WARNING: IEEEtran.bst: No hyphenation pattern has been}%
\typeout{** loaded for the language `#1'. Using the pattern for}%
\typeout{** the default language instead.}%
\else
\language=\csname l@#1\endcsname
\fi
#2}}
\providecommand{\BIBdecl}{\relax}
\BIBdecl

\bibitem{791823}
C.~Viard-Gaudin, P.~M. Lallican, S.~Knerr, and P.~Binter, ``The ireste on/off
  (ironoff) dual handwriting database,'' in \emph{Document Analysis and
  Recognition, 1999. ICDAR '99. Proceedings of the Fifth International
  Conference on}, Sep 1999, pp. 455--458.

\bibitem{Goodfellow-et-al-2016}
I.~Goodfellow, Y.~Bengio, and A.~Courville, \emph{Deep Learning}.\hskip 1em
  plus 0.5em minus 0.4em\relax MIT Press, 2016,
  \url{http://www.deeplearningbook.org}.

\bibitem{kingma2013auto}
D.~P. Kingma and M.~Welling, ``Auto-encoding variational bayes,'' \emph{arXiv
  preprint arXiv:1312.6114}, 2013.

\bibitem{goodfellow2014generative}
I.~Goodfellow, J.~Pouget-Abadie, M.~Mirza, B.~Xu, D.~Warde-Farley, S.~Ozair,
  A.~Courville, and Y.~Bengio, ``Generative adversarial nets,'' in
  \emph{Advances in neural information processing systems}, 2014, pp.
  2672--2680.

\bibitem{hochreiter1997long}
S.~Hochreiter and J.~Schmidhuber, ``Long short-term memory,'' \emph{Neural
  computation}, vol.~9, no.~8, pp. 1735--1780, 1997.

\bibitem{cho2014learning}
K.~Cho, B.~Van~Merri{\"e}nboer, C.~Gulcehre, D.~Bahdanau, F.~Bougares,
  H.~Schwenk, and Y.~Bengio, ``Learning phrase representations using rnn
  encoder-decoder for statistical machine translation,'' \emph{arXiv preprint
  arXiv:1406.1078}, 2014.

\bibitem{chung2014empirical}
J.~Chung, C.~Gulcehre, K.~Cho, and Y.~Bengio, ``Empirical evaluation of gated
  recurrent neural networks on sequence modeling,'' \emph{arXiv preprint
  arXiv:1412.3555}, 2014.

\bibitem{sutskever2011generating}
I.~Sutskever, J.~Martens, and G.~E. Hinton, ``Generating text with recurrent
  neural networks,'' in \emph{Proceedings of the 28th International Conference
  on Machine Learning (ICML-11)}, 2011, pp. 1017--1024.

\bibitem{Sutskever:2014:SSL:2969033.2969173}
\BIBentryALTinterwordspacing
I.~Sutskever, O.~Vinyals, and Q.~V. Le, ``Sequence to sequence learning with
  neural networks,'' in \emph{Proceedings of the 27th International Conference
  on Neural Information Processing Systems - Volume 2}, ser. NIPS'14.\hskip 1em
  plus 0.5em minus 0.4em\relax Cambridge, MA, USA: MIT Press, 2014, pp.
  3104--3112. [Online]. Available:
  \url{http://dl.acm.org/citation.cfm?id=2969033.2969173}
\BIBentrySTDinterwordspacing

\bibitem{karpathy2015deep}
A.~Karpathy and L.~Fei-Fei, ``Deep visual-semantic alignments for generating
  image descriptions,'' in \emph{Proceedings of the IEEE conference on computer
  vision and pattern recognition}, 2015, pp. 3128--3137.

\bibitem{vinyals2015show}
O.~Vinyals, A.~Toshev, S.~Bengio, and D.~Erhan, ``Show and tell: A neural image
  caption generator,'' in \emph{Computer Vision and Pattern Recognition (CVPR),
  2015 IEEE Conference on}.\hskip 1em plus 0.5em minus 0.4em\relax IEEE, 2015,
  pp. 3156--3164.

\bibitem{briot2017music}
J.-P. Briot and F.~Pachet, ``Music generation by deep learning-challenges and
  directions,'' \emph{arXiv preprint arXiv:1712.04371}, 2017.

\bibitem{oord2016wavenet}
A.~v.~d. Oord, S.~Dieleman, H.~Zen, K.~Simonyan, O.~Vinyals, A.~Graves,
  N.~Kalchbrenner, A.~Senior, and K.~Kavukcuoglu, ``Wavenet: A generative model
  for raw audio,'' \emph{arXiv preprint arXiv:1609.03499}, 2016.

\bibitem{DBLP:journals/corr/Graves13}
\BIBentryALTinterwordspacing
A.~Graves, ``Generating sequences with recurrent neural networks,''
  \emph{CoRR}, vol. abs/1308.0850, 2013. [Online]. Available:
  \url{http://arxiv.org/abs/1308.0850}
\BIBentrySTDinterwordspacing

\bibitem{bishop1994mixture}
C.~M. Bishop, ``Mixture density networks,'' 1994.

\bibitem{marti1999full}
U.-V. Marti and H.~Bunke, ``A full english sentence database for off-line
  handwriting recognition,'' in \emph{Document Analysis and Recognition, 1999.
  ICDAR'99. Proceedings of the Fifth International Conference on}.\hskip 1em
  plus 0.5em minus 0.4em\relax IEEE, 1999, pp. 705--708.

\bibitem{Theis:2015:GIM:2969442.2969455}
\BIBentryALTinterwordspacing
L.~Theis and M.~Bethge, ``Generative image modeling using spatial lstms,'' in
  \emph{Proceedings of the 28th International Conference on Neural Information
  Processing Systems - Volume 2}, ser. NIPS'15.\hskip 1em plus 0.5em minus
  0.4em\relax Cambridge, MA, USA: MIT Press, 2015, pp. 1927--1935. [Online].
  Available: \url{http://dl.acm.org/citation.cfm?id=2969442.2969455}
\BIBentrySTDinterwordspacing

\bibitem{VanDenOord:2016:PRN:3045390.3045575}
\BIBentryALTinterwordspacing
A.~Van Den~Oord, N.~Kalchbrenner, and K.~Kavukcuoglu, ``Pixel recurrent neural
  networks,'' in \emph{Proceedings of the 33rd International Conference on
  International Conference on Machine Learning - Volume 48}, ser.
  ICML'16.\hskip 1em plus 0.5em minus 0.4em\relax JMLR.org, 2016, pp.
  1747--1756. [Online]. Available:
  \url{http://dl.acm.org/citation.cfm?id=3045390.3045575}
\BIBentrySTDinterwordspacing

\bibitem{DBLP:journals/corr/abs-1803-09047}
\BIBentryALTinterwordspacing
R.~J. Skerry{-}Ryan, E.~Battenberg, Y.~Xiao, Y.~Wang, D.~Stanton, J.~Shor,
  R.~J. Weiss, R.~Clark, and R.~A. Saurous, ``Towards end-to-end prosody
  transfer for expressive speech synthesis with tacotron,'' \emph{CoRR}, vol.
  abs/1803.09047, 2018. [Online]. Available:
  \url{http://arxiv.org/abs/1803.09047}
\BIBentrySTDinterwordspacing

\bibitem{DBLP:journals/corr/abs-1803-09017}
\BIBentryALTinterwordspacing
Y.~Wang, D.~Stanton, Y.~Zhang, R.~J. Skerry{-}Ryan, E.~Battenberg, J.~Shor,
  Y.~Xiao, F.~Ren, Y.~Jia, and R.~A. Saurous, ``Style tokens: Unsupervised
  style modeling, control and transfer in end-to-end speech synthesis,''
  \emph{CoRR}, vol. abs/1803.09017, 2018. [Online]. Available:
  \url{http://arxiv.org/abs/1803.09017}
\BIBentrySTDinterwordspacing

\bibitem{Koehn:2010:SMT:1734086}
P.~Koehn, \emph{Statistical Machine Translation}, 1st~ed.\hskip 1em plus 0.5em
  minus 0.4em\relax New York, NY, USA: Cambridge University Press, 2010.

\bibitem{papineni2002bleu}
K.~Papineni, S.~Roukos, T.~Ward, and W.-J. Zhu, ``Bleu: a method for automatic
  evaluation of machine translation,'' in \emph{Proceedings of the 40th annual
  meeting on association for computational linguistics}.\hskip 1em plus 0.5em
  minus 0.4em\relax Association for Computational Linguistics, 2002, pp.
  311--318.

\bibitem{banerjee2005meteor}
S.~Banerjee and A.~Lavie, ``Meteor: An automatic metric for mt evaluation with
  improved correlation with human judgments,'' in \emph{Proceedings of the acl
  workshop on intrinsic and extrinsic evaluation measures for machine
  translation and/or summarization}, 2005, pp. 65--72.

\bibitem{vedantam2015cider}
R.~Vedantam, C.~Lawrence~Zitnick, and D.~Parikh, ``Cider: Consensus-based image
  description evaluation,'' in \emph{Proceedings of the IEEE conference on
  computer vision and pattern recognition}, 2015, pp. 4566--4575.

\bibitem{freeman1961encoding}
H.~Freeman, ``On the encoding of arbitrary geometric configurations,''
  \emph{IRE Transactions on Electronic Computers}, vol.~2, pp. 260--268, 1961.

\bibitem{kingma2014adam}
D.~P. Kingma and J.~Ba, ``Adam: A method for stochastic optimization,''
  \emph{arXiv preprint arXiv:1412.6980}, 2014.

\bibitem{Bengio:2015:SSS:2969239.2969370}
\BIBentryALTinterwordspacing
S.~Bengio, O.~Vinyals, N.~Jaitly, and N.~Shazeer, ``Scheduled sampling for
  sequence prediction with recurrent neural networks,'' in \emph{Proceedings of
  the 28th International Conference on Neural Information Processing Systems -
  Volume 1}, ser. NIPS'15.\hskip 1em plus 0.5em minus 0.4em\relax Cambridge,
  MA, USA: MIT Press, 2015, pp. 1171--1179. [Online]. Available:
  \url{http://dl.acm.org/citation.cfm?id=2969239.2969370}
\BIBentrySTDinterwordspacing

\bibitem{nguyen2010techniques}
V.~Nguyen and M.~Blumenstein, ``Techniques for static handwriting trajectory
  recovery: a survey,'' in \emph{Proceedings of the 9th IAPR International
  Workshop on Document Analysis Systems}.\hskip 1em plus 0.5em minus
  0.4em\relax ACM, 2010, pp. 463--470.

\bibitem{10.2307/3001968}
\BIBentryALTinterwordspacing
F.~Wilcoxon, ``Individual comparisons by ranking methods,'' \emph{Biometrics
  Bulletin}, vol.~1, no.~6, pp. 80--83, 1945. [Online]. Available:
  \url{http://www.jstor.org/stable/3001968}
\BIBentrySTDinterwordspacing

\bibitem{DBLP:journals/corr/SchroffKP15}
\BIBentryALTinterwordspacing
F.~Schroff, D.~Kalenichenko, and J.~Philbin, ``Facenet: {A} unified embedding
  for face recognition and clustering,'' \emph{CoRR}, vol. abs/1503.03832,
  2015. [Online]. Available: \url{http://arxiv.org/abs/1503.03832}
\BIBentrySTDinterwordspacing

\bibitem{DBLP:journals/corr/Bredin16}
\BIBentryALTinterwordspacing
H.~Bredin, ``Tristounet: Triplet loss for speaker turn embedding,''
  \emph{CoRR}, vol. abs/1609.04301, 2016. [Online]. Available:
  \url{http://arxiv.org/abs/1609.04301}
\BIBentrySTDinterwordspacing

\end{thebibliography}

\clearpage

\appendix
\section*{Example of the letters}
\par
\begin{figure*}
\centering
    \begin{subfigure}[b]{0.10\textwidth}
        \includegraphics[width=\textwidth]{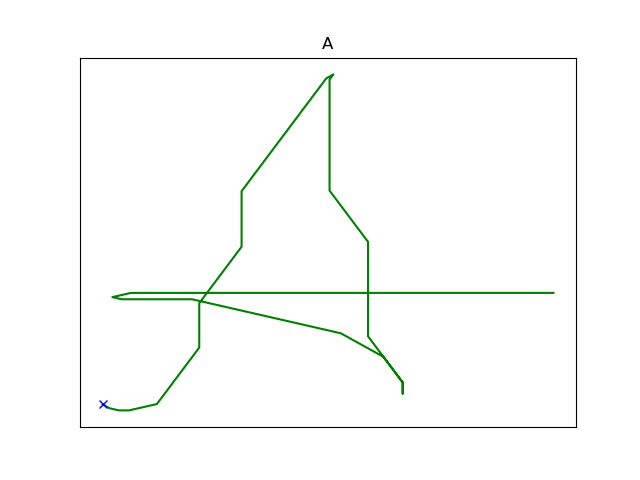}
        \caption{B}
    \end{subfigure}
    ~
    \begin{subfigure}[b]{0.10\textwidth}
        \includegraphics[width=\textwidth]{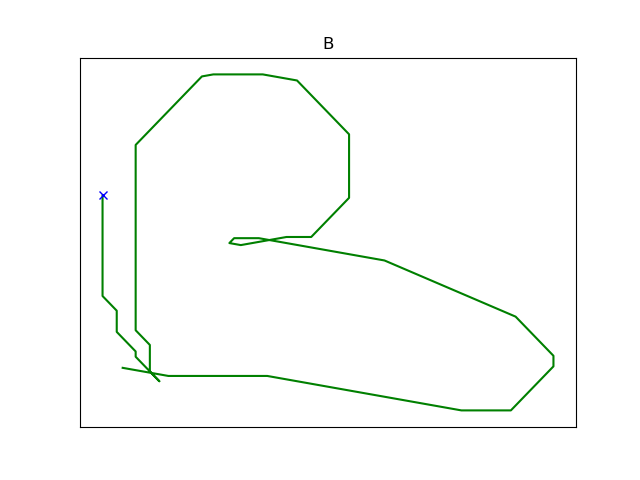}
        \caption{C}
    \end{subfigure}
    ~
    \begin{subfigure}[b]{0.10\textwidth}
        \includegraphics[width=\textwidth]{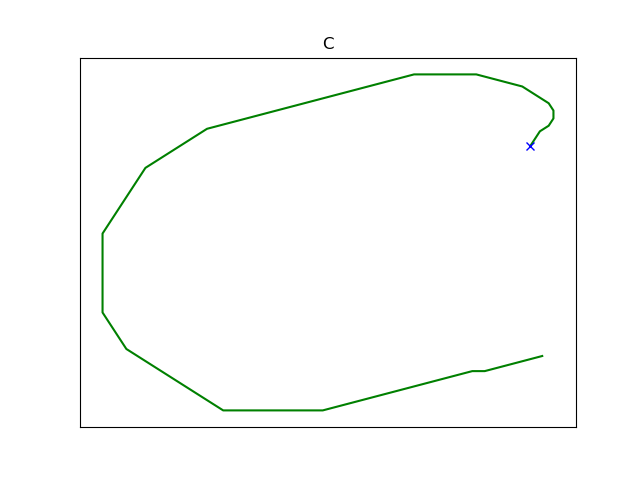}
        \caption{D}
    \end{subfigure}
    ~
    \begin{subfigure}[b]{0.10\textwidth}
        \includegraphics[width=\textwidth]{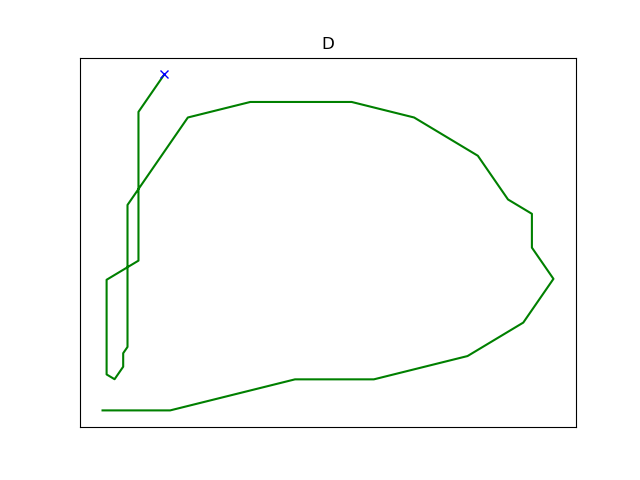}
        \caption{A}
    \end{subfigure}
    ~ 
    \begin{subfigure}[b]{0.10\textwidth}
        \includegraphics[width=\textwidth]{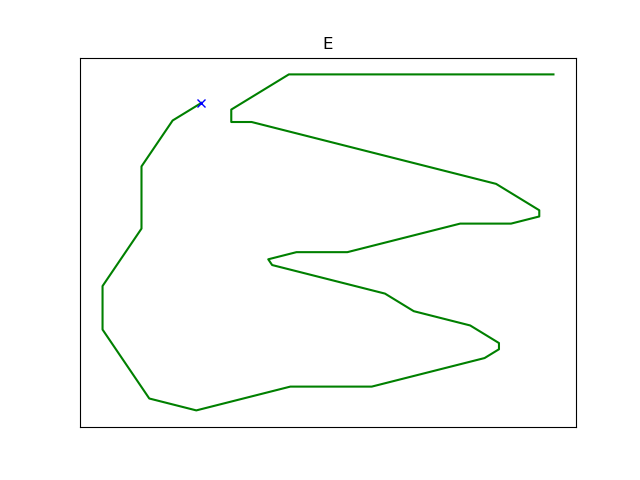}
        \caption{E}
    \end{subfigure}
    ~
    \begin{subfigure}[b]{0.10\textwidth}
        \includegraphics[width=\textwidth]{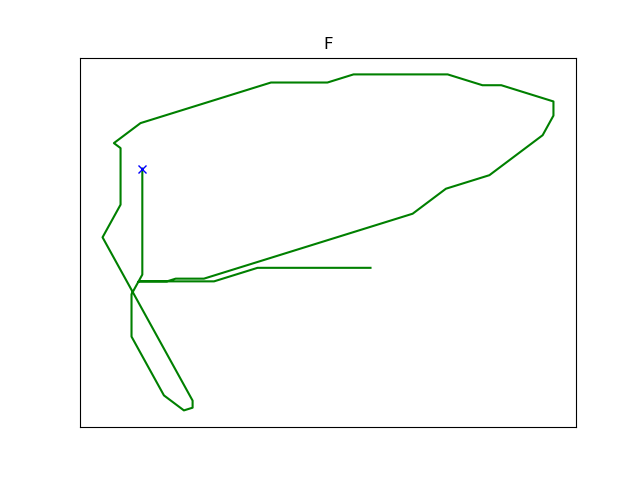}
        \caption{F}
    \end{subfigure}
    ~
    \begin{subfigure}[b]{0.10\textwidth}
        \includegraphics[width=\textwidth]{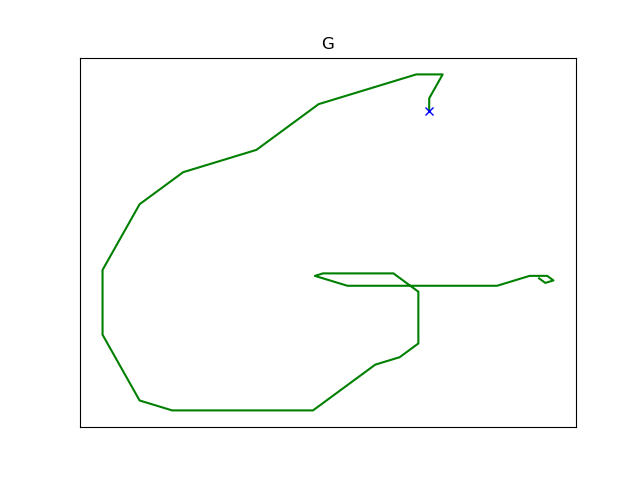}
        \caption{G}
    \end{subfigure}
    ~
    \begin{subfigure}[b]{0.10\textwidth}
        \includegraphics[width=\textwidth]{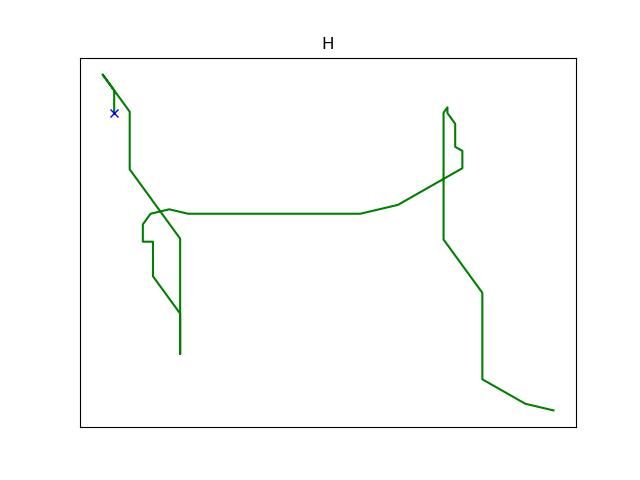}
        \caption{H}
    \end{subfigure}
    ~
    \begin{subfigure}[b]{0.10\textwidth}
        \includegraphics[width=\textwidth]{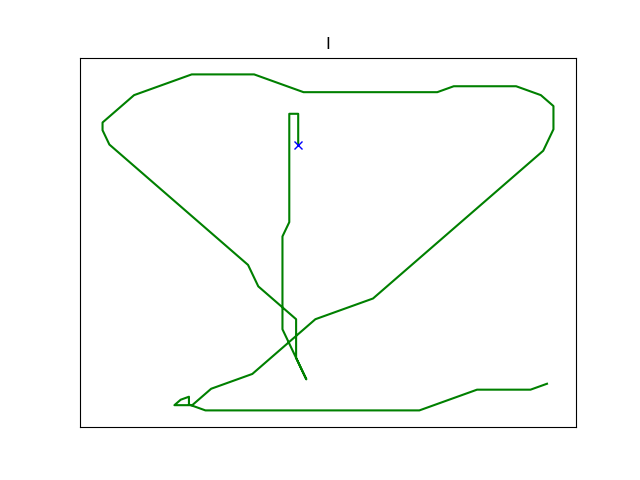}
        \caption{I}
    \end{subfigure}
    ~
    \begin{subfigure}[b]{0.10\textwidth}
        \includegraphics[width=\textwidth]{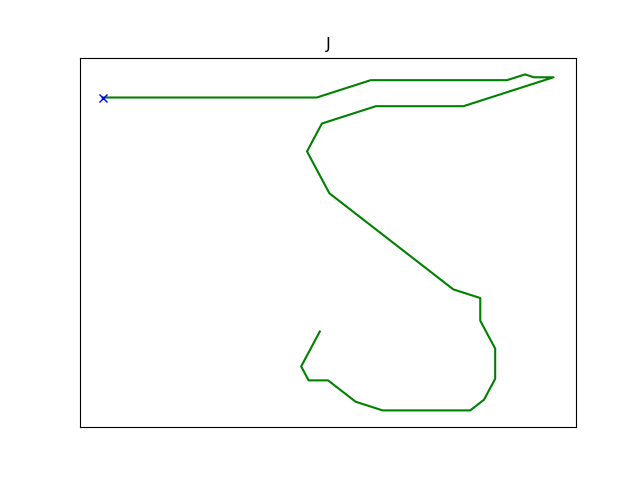}
        \caption{J}
    \end{subfigure}
    ~
    \begin{subfigure}[b]{0.10\textwidth}
        \includegraphics[width=\textwidth]{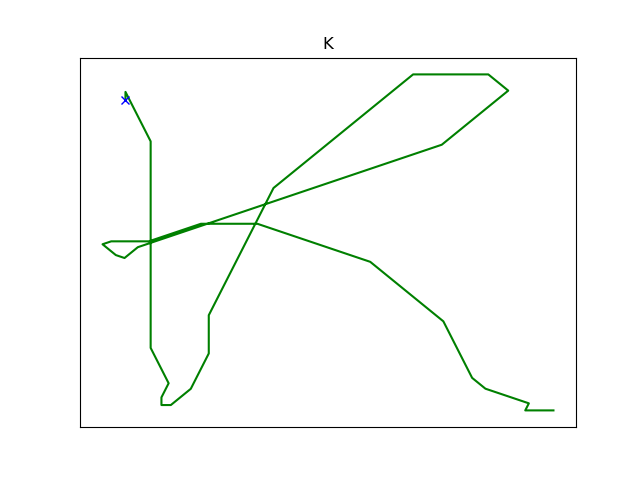}
        \caption{K}
    \end{subfigure}
    ~
    \begin{subfigure}[b]{0.10\textwidth}
        \includegraphics[width=\textwidth]{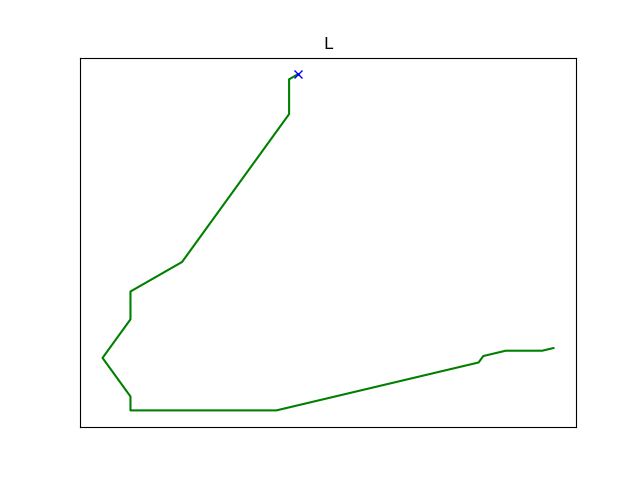}
        \caption{L}
    \end{subfigure}
    ~
    \begin{subfigure}[b]{0.10\textwidth}
        \includegraphics[width=\textwidth]{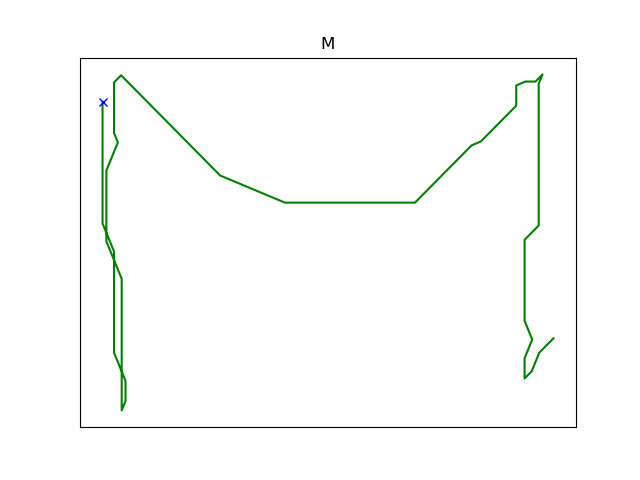}
        \caption{M}
    \end{subfigure}
    ~
    \begin{subfigure}[b]{0.10\textwidth}
        \includegraphics[width=\textwidth]{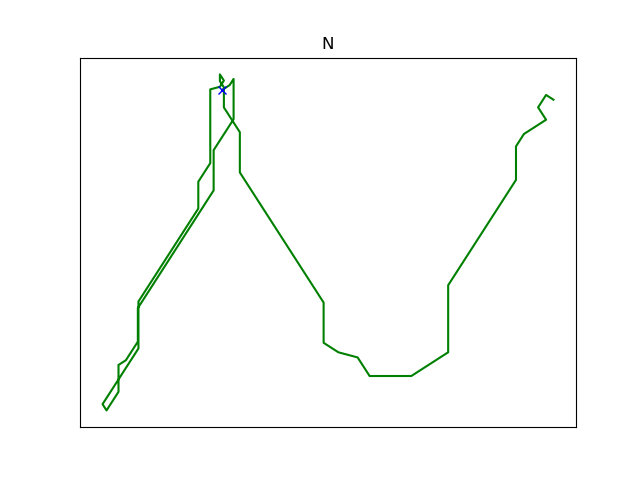}
        \caption{N}
    \end{subfigure}
    ~
    \begin{subfigure}[b]{0.10\textwidth}
        \includegraphics[width=\textwidth]{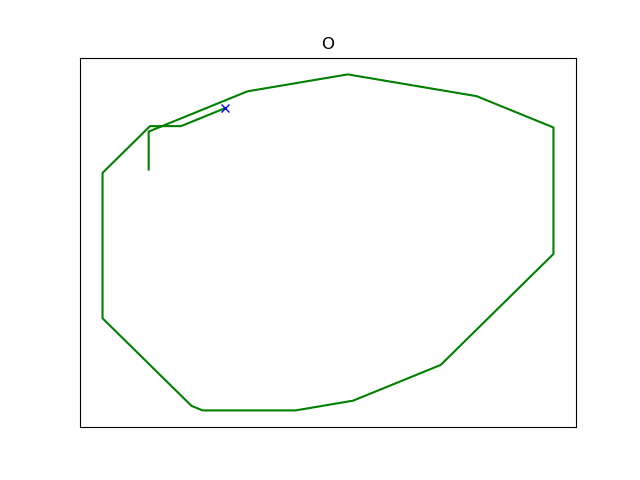}
        \caption{O}
    \end{subfigure}
    ~
    \begin{subfigure}[b]{0.10\textwidth}
        \includegraphics[width=\textwidth]{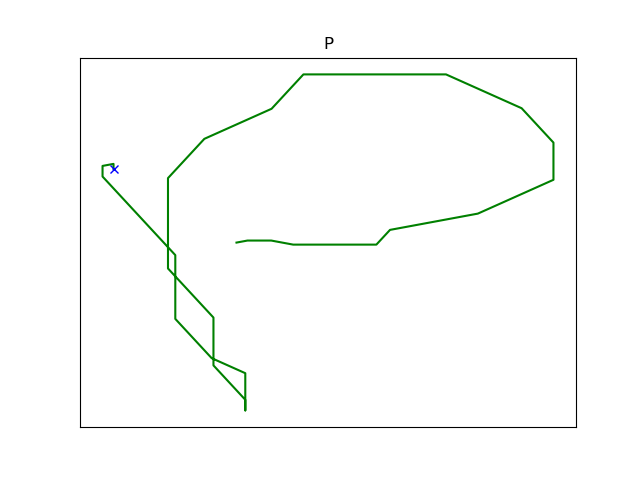}
        \caption{P}
    \end{subfigure}
    ~
    \begin{subfigure}[b]{0.10\textwidth}
        \includegraphics[width=\textwidth]{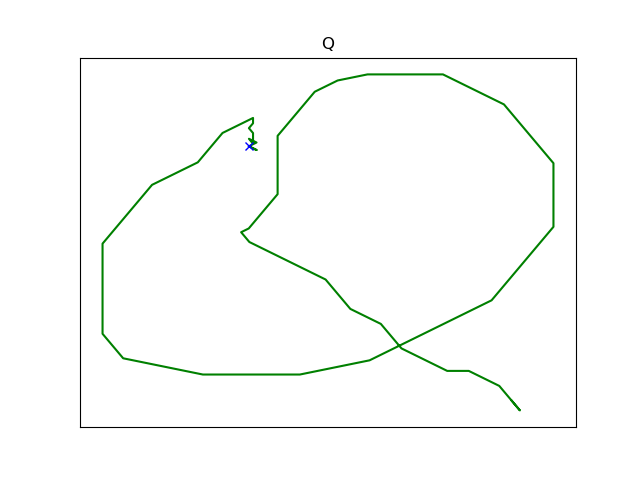}
        \caption{Q}
    \end{subfigure}
    ~
    \begin{subfigure}[b]{0.10\textwidth}
        \includegraphics[width=\textwidth]{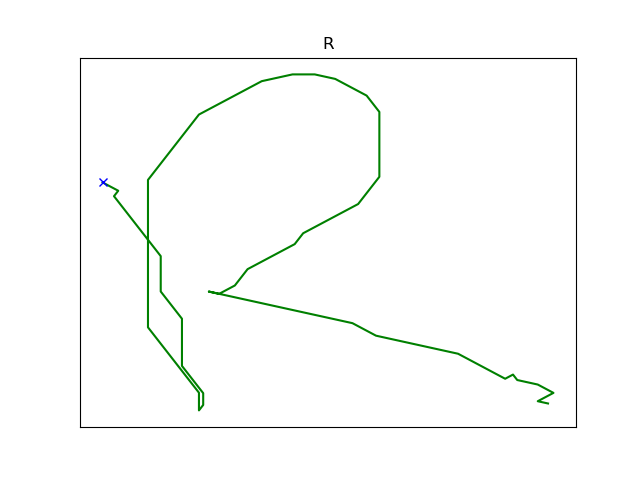}
        \caption{R}
    \end{subfigure}
    ~
    \begin{subfigure}[b]{0.10\textwidth}
        \includegraphics[width=\textwidth]{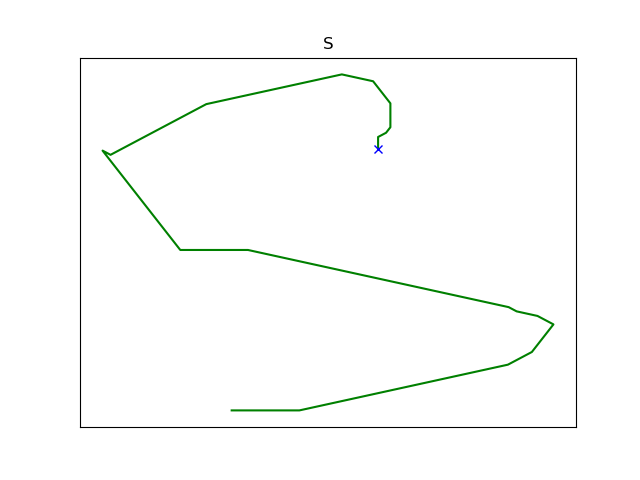}
        \caption{S}
    \end{subfigure}
    ~
    \begin{subfigure}[b]{0.10\textwidth}
        \includegraphics[width=\textwidth]{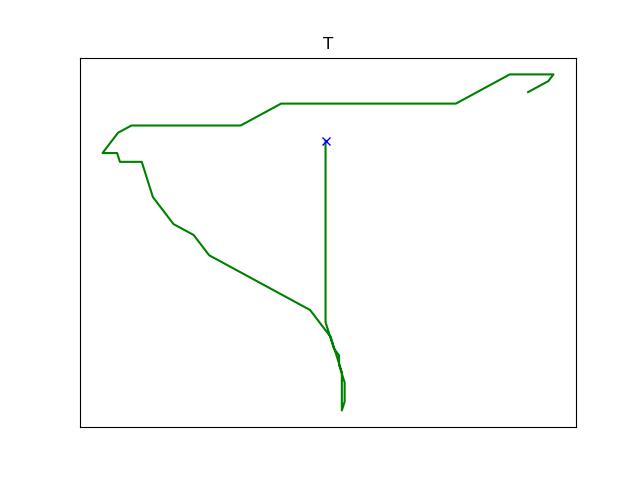}
        \caption{T}
    \end{subfigure}
    ~
    \begin{subfigure}[b]{0.10\textwidth}
        \includegraphics[width=\textwidth]{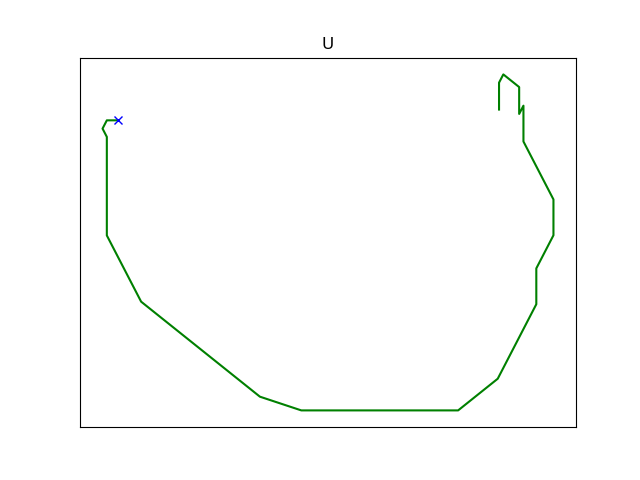}
        \caption{U}
    \end{subfigure}
    ~
    \begin{subfigure}[b]{0.10\textwidth}
        \includegraphics[width=\textwidth]{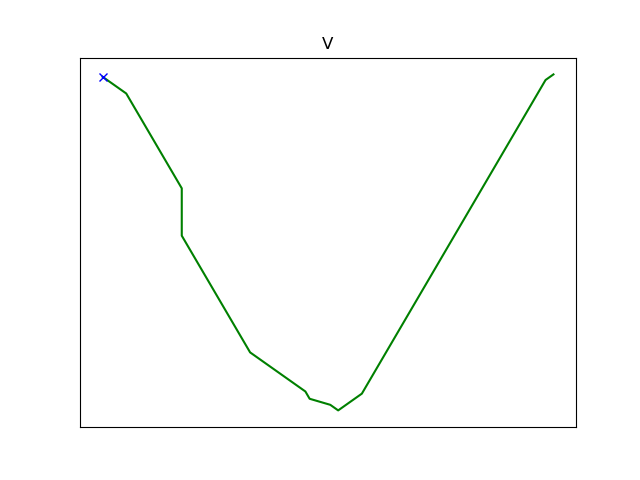}
        \caption{V}
    \end{subfigure}
    ~
    \begin{subfigure}[b]{0.10\textwidth}
        \includegraphics[width=\textwidth]{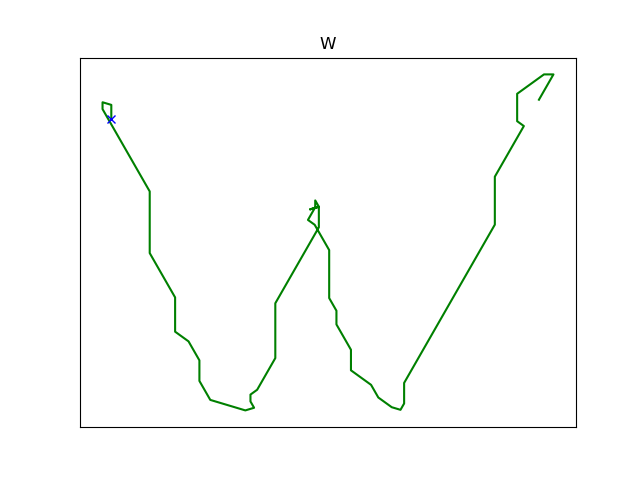}
        \caption{W}
    \end{subfigure}
    ~
    \begin{subfigure}[b]{0.10\textwidth}
        \includegraphics[width=\textwidth]{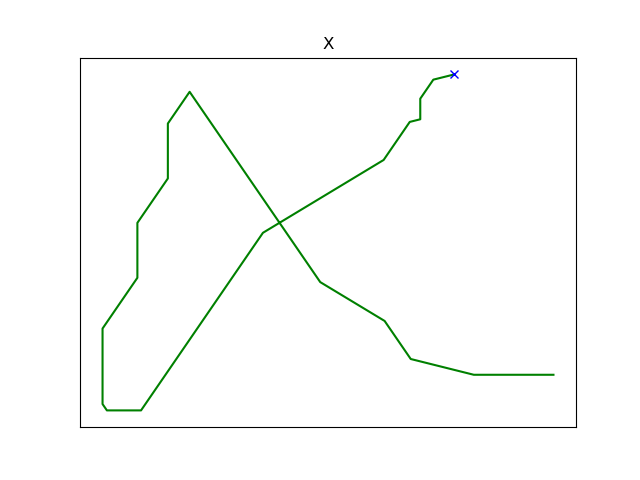}
        \caption{X}
    \end{subfigure}
    ~
    \begin{subfigure}[b]{0.10\textwidth}
        \includegraphics[width=\textwidth]{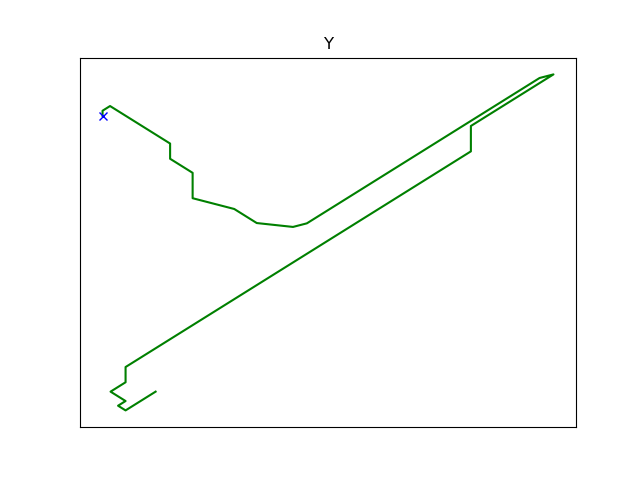}
        \caption{Y}
    \end{subfigure}
    ~
    \begin{subfigure}[b]{0.10\textwidth}
        \includegraphics[width=\textwidth]{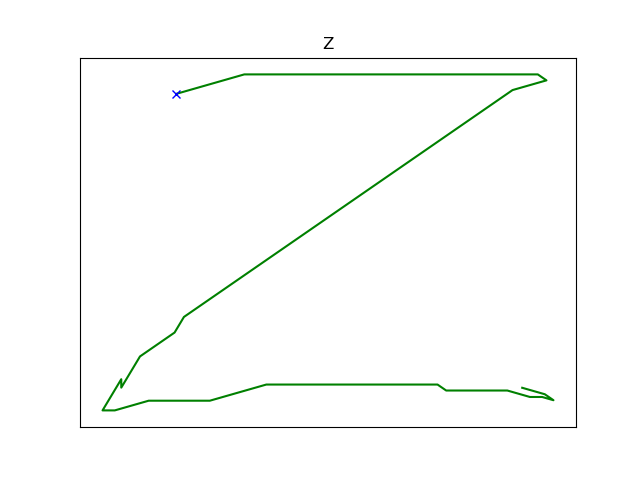}
        \caption{Z}
    \end{subfigure}
    
    \caption{Examples of original letters. The blue \textit{x} mark is the starting point. These ones are generated using the letter + Writer bias. E and F are visually harder to recognize, since we do not model the pen pressure, otherwise, the rest of the letters are well recognizable.}\label{fig:orig_letters_examples}
\end{figure*}

\begin{figure*}[!htbp]
\centering
    \begin{subfigure}[b]{0.10\textwidth}
        \includegraphics[width=\textwidth]{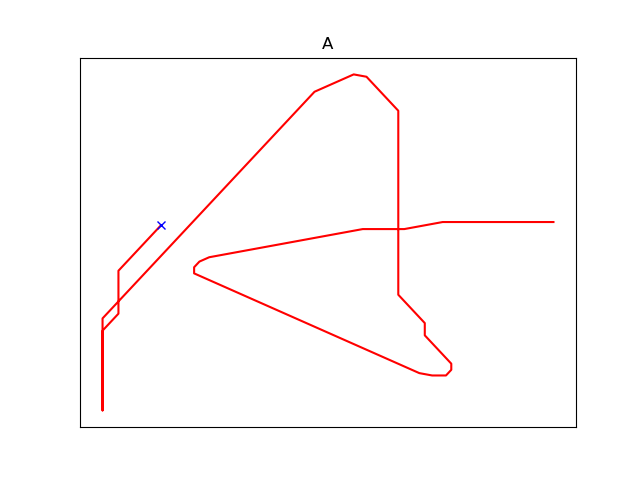}
        \caption{A}
    \end{subfigure}
    ~ 
    \begin{subfigure}[b]{0.10\textwidth}
        \includegraphics[width=\textwidth]{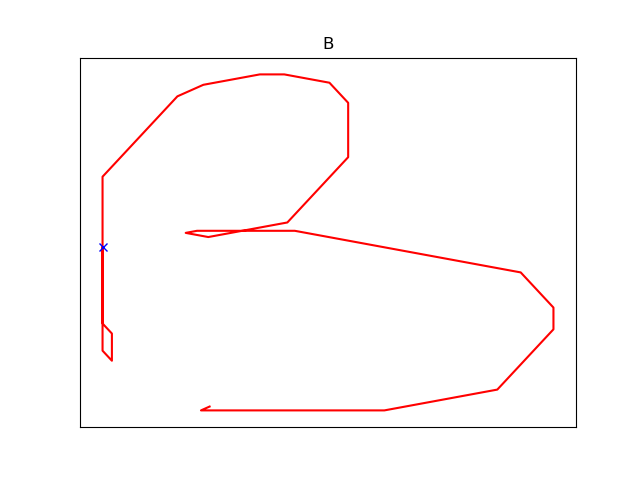}
        \caption{B}
    \end{subfigure}
    ~
    \begin{subfigure}[b]{0.10\textwidth}
        \includegraphics[width=\textwidth]{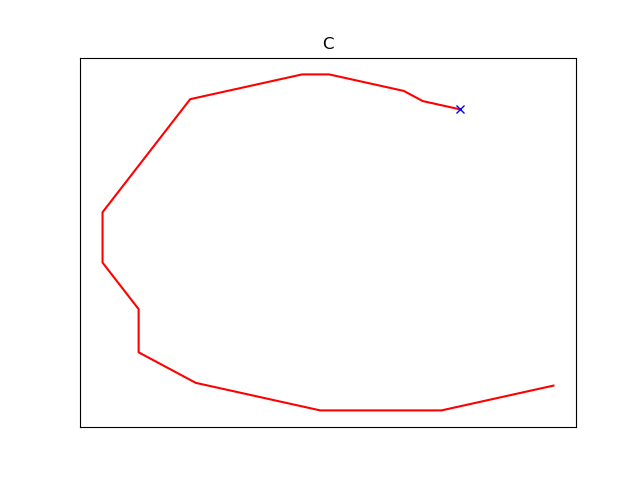}
        \caption{C}
    \end{subfigure}
    ~
    \begin{subfigure}[b]{0.10\textwidth}
        \includegraphics[width=\textwidth]{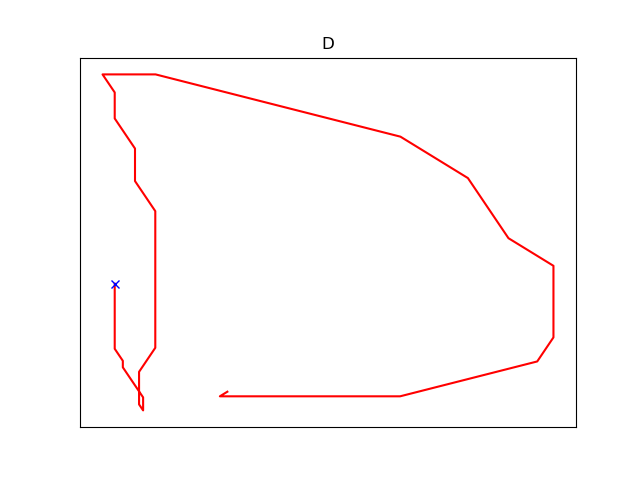}
        \caption{D}
    \end{subfigure}
    ~
    \begin{subfigure}[b]{0.10\textwidth}
        \includegraphics[width=\textwidth]{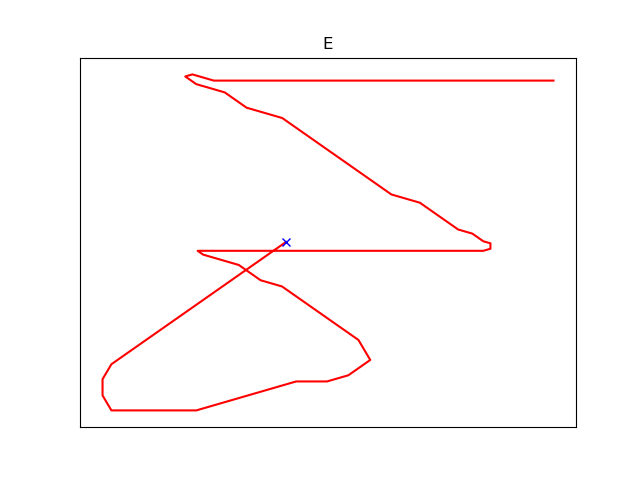}
        \caption{E}
    \end{subfigure}
    ~
    \begin{subfigure}[b]{0.10\textwidth}
        \includegraphics[width=\textwidth]{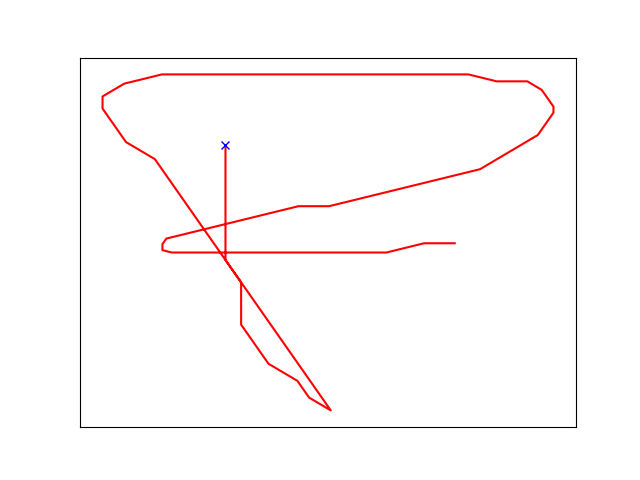}
        \caption{F}
    \end{subfigure}
    ~
    \begin{subfigure}[b]{0.10\textwidth}
        \includegraphics[width=\textwidth]{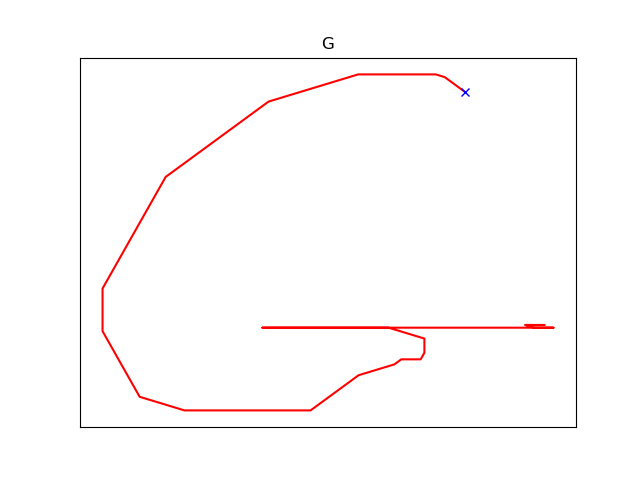}
        \caption{G}
    \end{subfigure}
    ~
    \begin{subfigure}[b]{0.10\textwidth}
        \includegraphics[width=\textwidth]{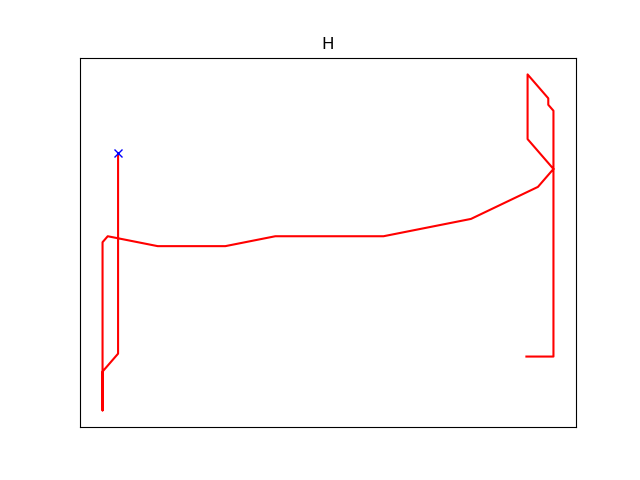}
        \caption{H}
    \end{subfigure}
    ~
    \begin{subfigure}[b]{0.10\textwidth}
        \includegraphics[width=\textwidth]{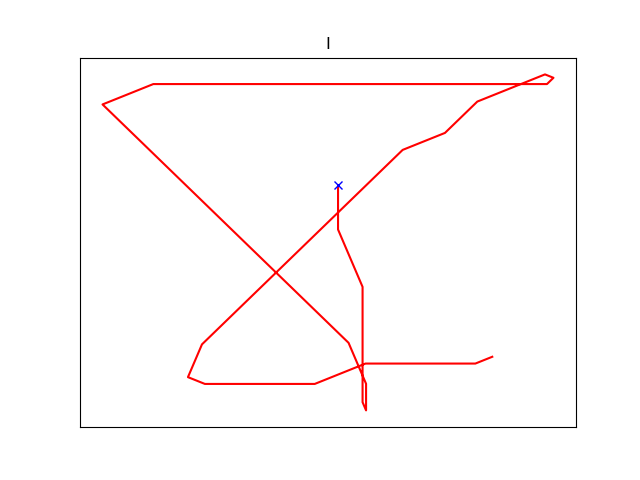}
        \caption{I}
    \end{subfigure}
    ~
    \begin{subfigure}[b]{0.10\textwidth}
        \includegraphics[width=\textwidth]{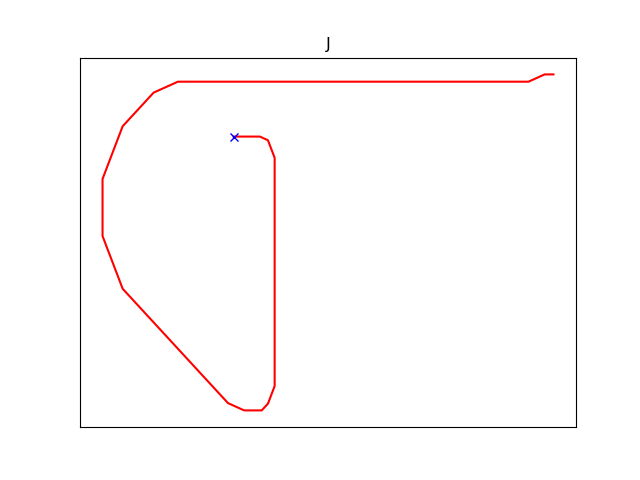}
        \caption{J}
    \end{subfigure}
    ~
    \begin{subfigure}[b]{0.10\textwidth}
        \includegraphics[width=\textwidth]{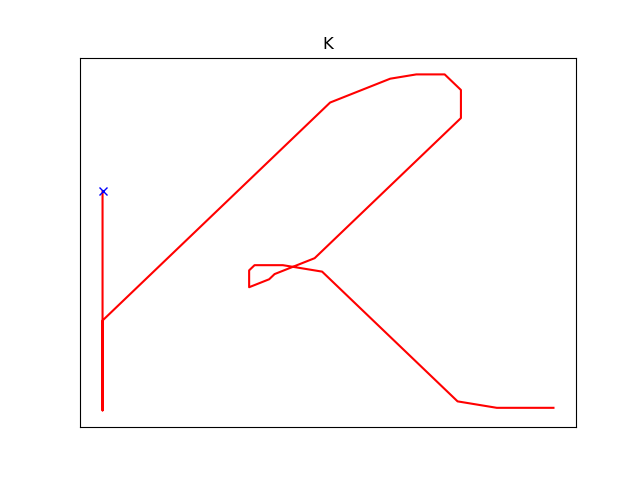}
        \caption{K}
    \end{subfigure}
    ~
    \begin{subfigure}[b]{0.10\textwidth}
        \includegraphics[width=\textwidth]{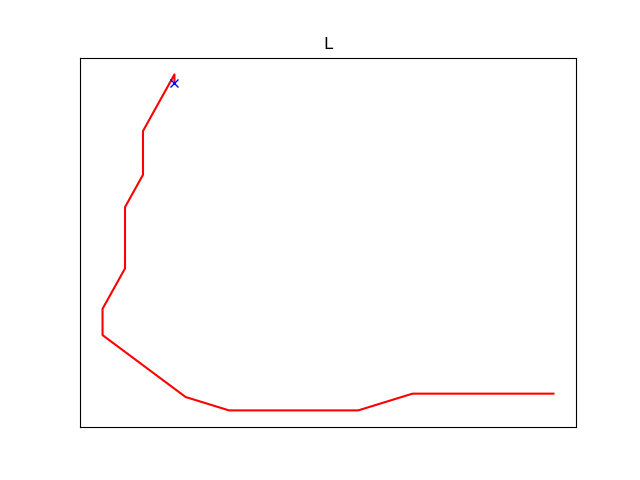}
        \caption{L}
    \end{subfigure}
    ~
    \begin{subfigure}[b]{0.10\textwidth}
        \includegraphics[width=\textwidth]{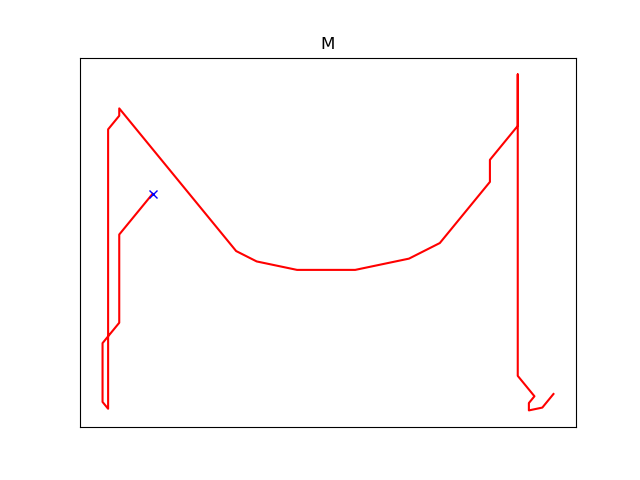}
        \caption{M}
    \end{subfigure}
    ~
    \begin{subfigure}[b]{0.10\textwidth}
        \includegraphics[width=\textwidth]{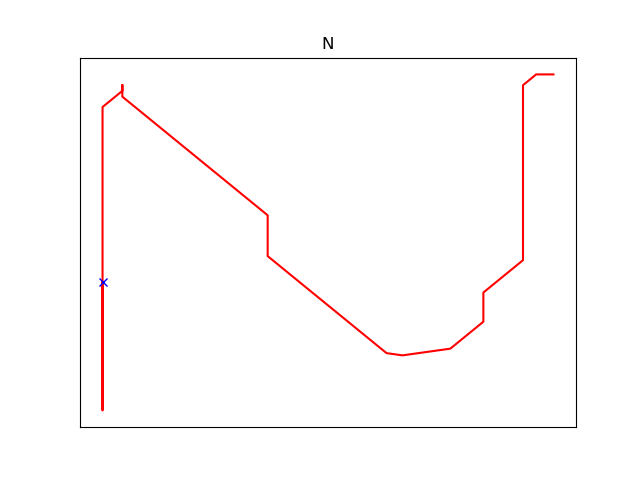}
        \caption{N}
    \end{subfigure}
    ~
    \begin{subfigure}[b]{0.10\textwidth}
        \includegraphics[width=\textwidth]{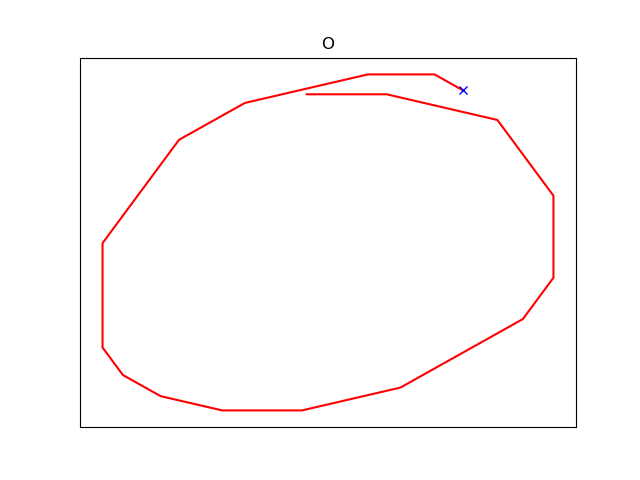}
        \caption{O}
    \end{subfigure}
    ~
    \begin{subfigure}[b]{0.10\textwidth}
        \includegraphics[width=\textwidth]{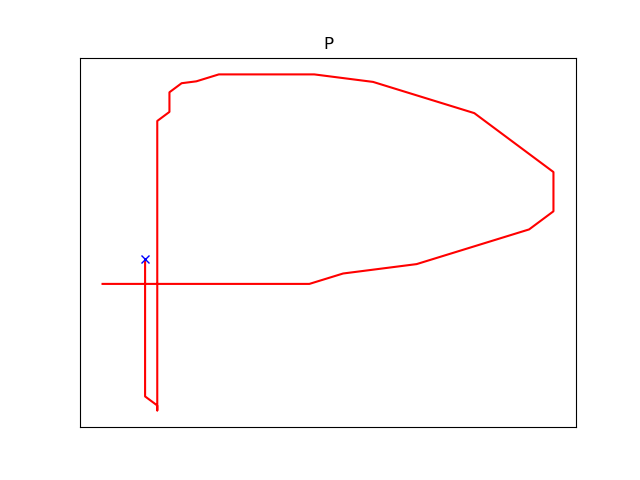}
        \caption{P}
    \end{subfigure}
    ~
    \begin{subfigure}[b]{0.10\textwidth}
        \includegraphics[width=\textwidth]{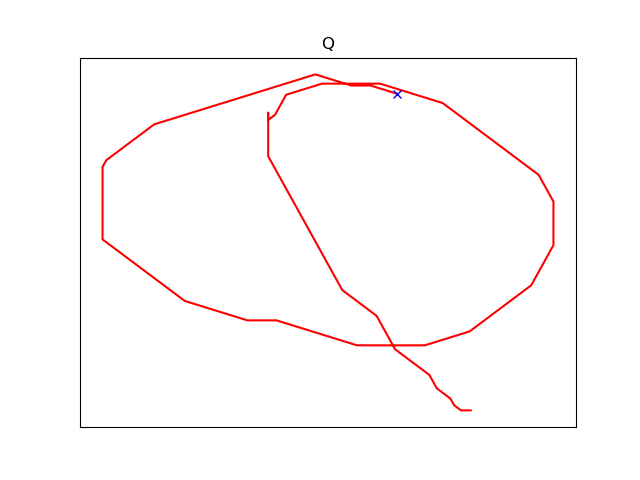}
        \caption{Q}
    \end{subfigure}
    ~
    \begin{subfigure}[b]{0.10\textwidth}
        \includegraphics[width=\textwidth]{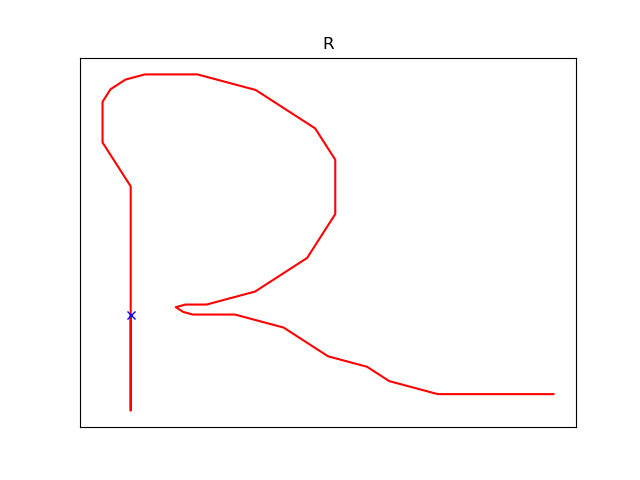}
        \caption{R}
    \end{subfigure}
    ~
    \begin{subfigure}[b]{0.10\textwidth}
        \includegraphics[width=\textwidth]{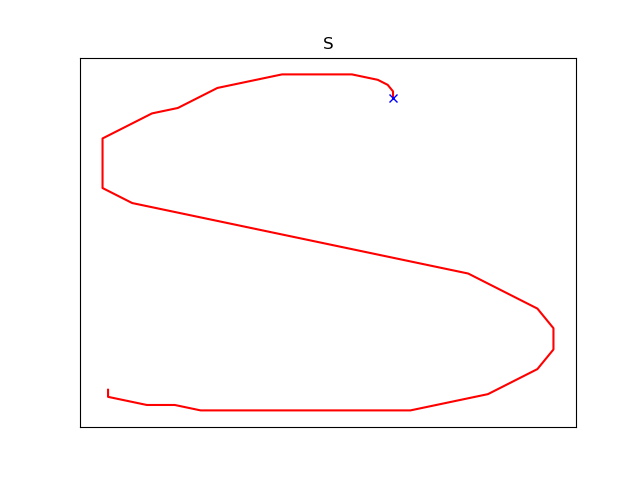}
        \caption{S}
    \end{subfigure}
    ~
    \begin{subfigure}[b]{0.10\textwidth}
        \includegraphics[width=\textwidth]{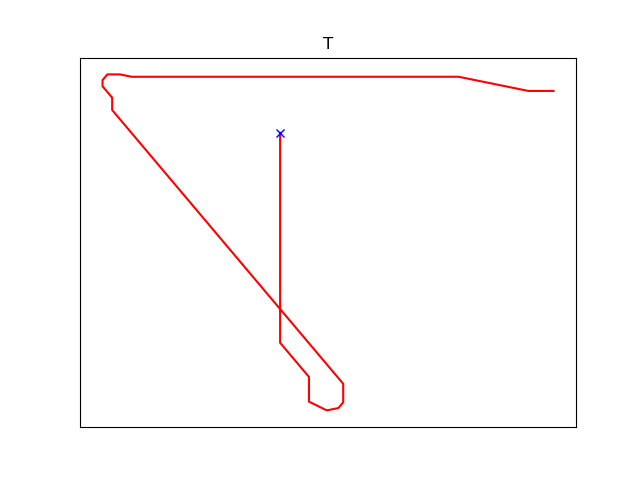}
        \caption{T}
    \end{subfigure}
    ~
    \begin{subfigure}[b]{0.10\textwidth}
        \includegraphics[width=\textwidth]{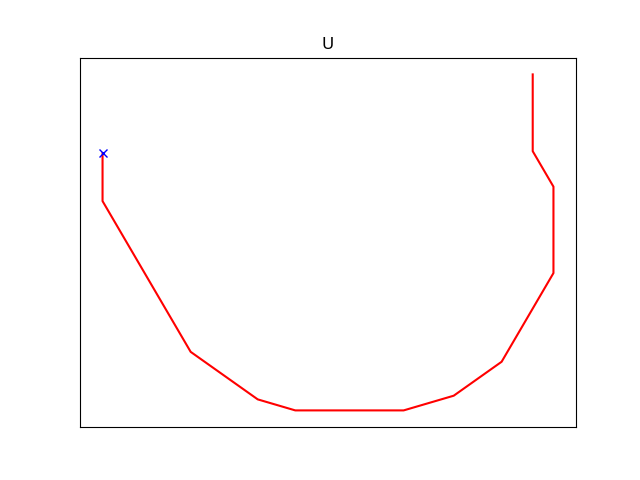}
        \caption{U}
    \end{subfigure}
    ~
    \begin{subfigure}[b]{0.10\textwidth}
        \includegraphics[width=\textwidth]{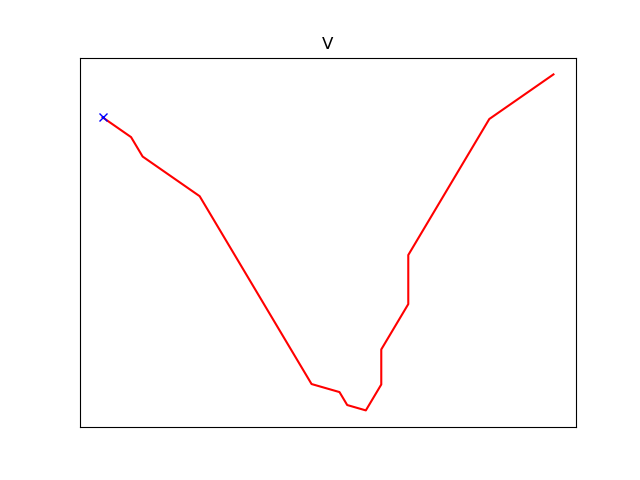}
        \caption{V}
    \end{subfigure}
    ~
    \begin{subfigure}[b]{0.10\textwidth}
        \includegraphics[width=\textwidth]{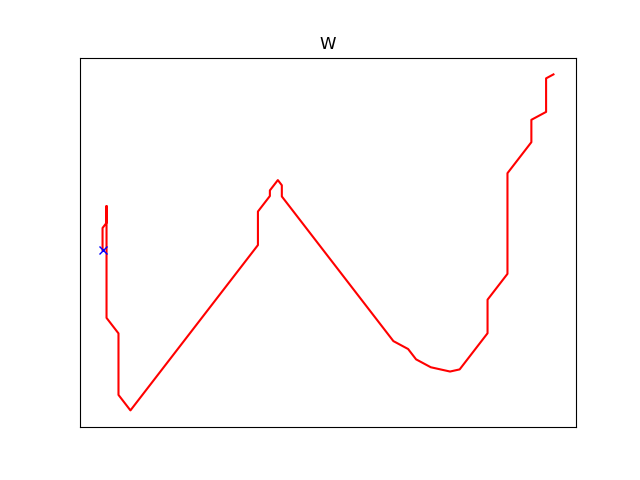}
        \caption{W}
    \end{subfigure}
    ~
    \begin{subfigure}[b]{0.10\textwidth}
        \includegraphics[width=\textwidth]{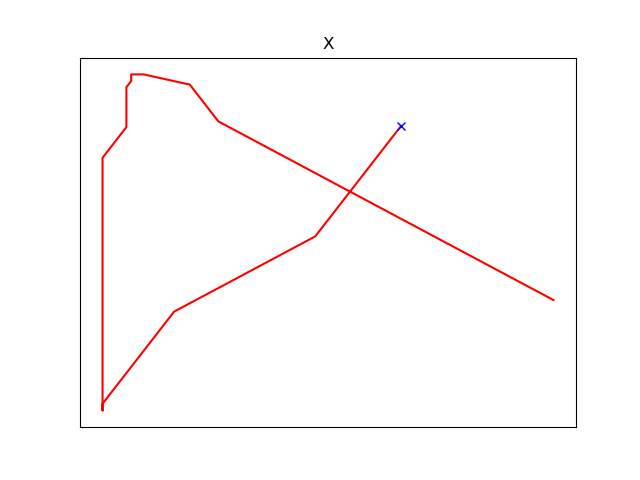}
        \caption{X}
    \end{subfigure}
    ~
    \begin{subfigure}[b]{0.10\textwidth}
        \includegraphics[width=\textwidth]{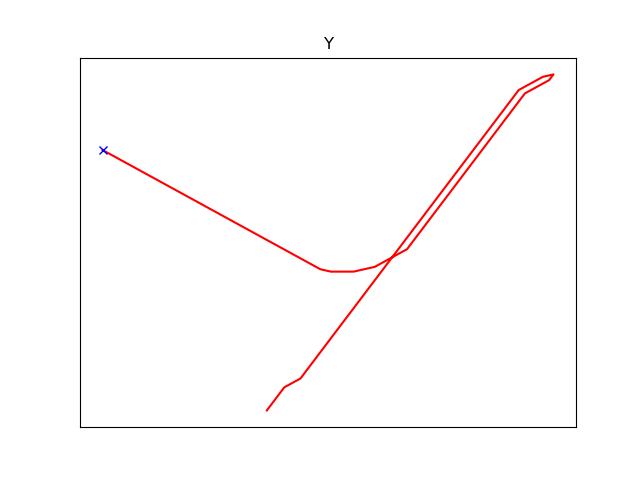}
        \caption{Y}
    \end{subfigure}
    ~
    \begin{subfigure}[b]{0.10\textwidth}
        \includegraphics[width=\textwidth]{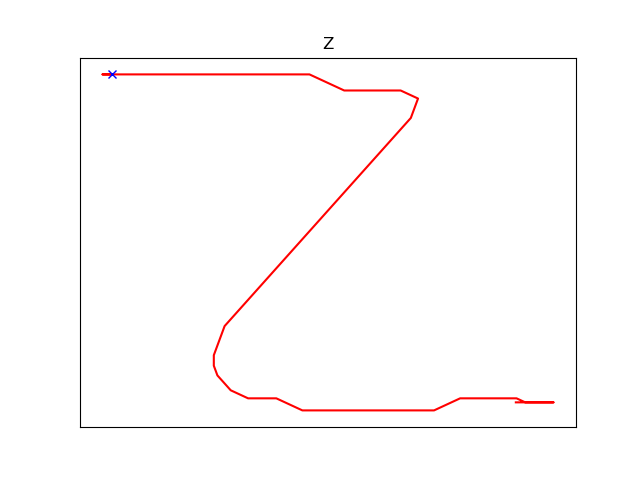}
        \caption{Z}
    \end{subfigure}
    
    \caption{Examples of generated letters. The blue \textit{x} mark is the starting point. These ones are generated using the letter + Writer bias. The general quality of this quite acceptable.}\label{fig:letters_examples}
\end{figure*}

 The design choices of our experiments (discretization, and ignoring the pen state) affects the final shape of the letters, yet, the letters and their style are quite recognizable. See examples for the original letters in figure \ref{fig:orig_letters_examples}. Examples for the generation with our methods are in figure \ref{fig:letters_examples}.

\end{document}